\documentclass[letterpaper]{article} 
\usepackage{aaai25}  
\usepackage{times}  
\usepackage{helvet}  
\usepackage{courier}  
\usepackage[hyphens]{url}  
\usepackage{graphicx} 
\usepackage{natbib}  
\usepackage{caption} 
\usepackage{algorithm}
\usepackage{algorithmic}

\usepackage{bbding}
\usepackage{amsmath,amsfonts}
\usepackage{tabularx}
\usepackage{tabulary}
\usepackage{booktabs}
\usepackage{multirow}
\usepackage{makecell}
\usepackage{cancel}
\usepackage[table, dvipsnames]{xcolor}
\newtheorem{theorem}{Theorem}

\newtheorem{assumption}{Assumption}

\usepackage{amssymb}
\definecolor{magiccolor}{RGB}{205, 232, 248}
\usepackage{newfloat}
\usepackage{listings}
\DeclareCaptionStyle{ruled}{labelfont=normalfont,labelsep=colon,strut=off} 
\floatstyle{ruled}
\newfloat{listing}{tb}{lst}{}
\floatname{listing}{Listing}
\title{MGDA: Model-based Goal Data Augmentation for Offline Goal-conditioned Weighted Supervised Learning}
\author {
    Xing Lei\textsuperscript{\rm 1}\quad
    Xuetao Zhang \textsuperscript{\rm 2}\thanks{Corresponding author}\quad
    Donglin Wang\textsuperscript{\rm 3}\quad
}
\affiliations {
    \textsuperscript{\rm 1,2}National Key Laboratory of Human-Machine Hybrid Augmented Intelligence, National Engineering Research Center for Visual Information and Applications, and Institute of Artificial Intelligence and Robotics, Xi'an Jiaotong University\\
    \textsuperscript{\rm 3}School of Enginneering, Westlake University\\
    leixing@stu.xjtu.edu.cn\quad 
    xuetaozhang@xjtu.edu.cn\quad
    wangdonglin@westlake.edu.cn
}

\usepackage{bibentry}

\begin{document}

\maketitle

\begin{abstract}
Recently, a state-of-the-art series of algorithms—Goal-Conditioned Weighted Supervised Learning (GCWSL) methods—has been introduced to address the challenges inherent in offline goal-conditioned reinforcement learning (RL). GCWSL optimizes a lower bound on the goal-conditioned RL objective and has demonstrated exceptional performance across a range of goal-reaching tasks, offering a simple, effective, and stable solution. Nonetheless, researches has revealed a critical limitation in GCWSL: the absence of trajectory stitching capabilities. In response, goal data augmentation strategies have been proposed to enhance these methods. However, existing techniques often fail to effectively sample appropriate augmented goals for GCWSL. In this paper, we establish unified principles for goal data augmentation, emphasizing goal diversity, action optimality, and goal reachability. Building on these principles, we propose a \underline{M}odel-based \underline{G}oal \underline{D}ata \underline{A}ugmentation (MGDA) approach, which leverages a dynamics model to sample more appropriate augmented goals. MGDA uniquely incorporates the local Lipschitz continuity assumption within the learned model to mitigate the effects of compounding errors. Empirical results demonstrate that MGDA significantly improves the performance of GCWSL methods on both state-based and vision-based maze datasets, outperforming previous goal data augmentation techniques in their ability to enhancing stitching capabilities.
\end{abstract}
\section{Introduction}
Deep reinforcement learning (RL) empowers agents to attain sophisticated objectives in complex and uncertain environments, such as computer games \citep{oroojlooyjadid2022deep, sestini2023deep}, robot control \citep{dargazany2021drl, quiroga2022position, plasencia2023deep}, and language processing \citep{akakzia2020grounding, sharifani2023machine}. One of the central challenges in deep RL is facilitating efficient learning in environments characterized by sparse rewards. This issue is particularly acute in goal-conditioned RL (GCRL)
\citep{kaelbling1993learning,schaul2015universal,andrychowicz2017hindsight,liu2022goal}, where the agent is tasked with learning generalized policies that can reach a variety of goals.
Offline GCRL is especially promising because it allows for the learning of goal-conditioned policies from purely offline datasets
without requiring any interaction with the environment during the learning process
\citep{levine2020offline}.
We note that some goal-conditioned weighted supervised learning (GCWSL) methods
\citep{yang2022rethinking,ma2022far,hejna2023distance,sikchi2024score}
were proposed to tackle the offline GCRL challenges. 
Compared with other goal-conditioned RL or self-supervised (SL) methods
\citep{yang2019imitation,srivastava2019training,chen2020learning,ding2019goal,lynch2020learning,paster2020planning,eysenbach2020c,ghosh2021learning,eysenbach2022contrastive},
GCWSL has demonstrated outstanding performance across various goal-reaching tasks in a simple, effective, and stable manner.

Although GCWSL has been successfully applied to effectively learn from sparse rewards in certain goal-reaching tasks within offline GCRL, 
some studies \citep{brandfonbrener2022does,yang2023swapped,ghugare2024closing} indicate that GCWSL may leads to sub-optimal policies when applied to the corresponding sub-trajectories of (state, goal) pairs across different trajectories and identify this issue as lacking of the ability to stitch trajectories. To tackle this problem, \citep{yang2023swapped} and \citep{ghugare2024closing} adopt goal data augmentation to sample more (state, goal) pairs during training phase.
Goal data augmentation is an effective data-driven approach that can generate diverse (state, goal) pairs. Therefore, it can enable GCWSL methods to exhibit stitching property, highlighting the strengths of supervised learning (SL). 

However,
our analysis indicates that they often fail to select appropriate goals as augmented goals.
This paper investigates this issue and proposes a more advanced goal data augmentation method.
Specifically, we establish three goal data augmentation unified principles, grounded in both the properties of data augmentation and the inherent characteristics of GCWSL : \textbf{\textit{Goal Diversity}}, \textbf{\textit{Action Optimality}}, and \textbf{\textit{Goal Reachability}}.
\textbf{\textit{Goal Diversity}} implies that a single initial state can correspond to multiple goals.
\textbf{\textit{Action Optimality}} implies that the action corresponding to a initial state should remain optimal for reaching the augmented goals. \textbf{\textit{Goal Reachability}} ensures that the augmented goals are reachable.
Building on these principles, 
we propose a model-based goal data augmentation (MGDA). 
MGDA leverages a dynamics model to predict nearby states around original goal, 
and then sample new augmented goals in the trajectory of these states.
To reduce the conformity error caused by model prediction,
we specifically employed the local Lipschitz continuity assumption when learning the dynamics model from the dataset.

We briefly summarize our contributions:
\begin{itemize}
	\item To the best of our knowledge, this work is the first to establish a set of principles specifically designed to enhance the stitching capabilities of GCWSL methods. Moreover, these principles are broadly applicable and can be extended to other goal-conditioned SL frameworks.
	\item We introduce a model-based data augmentation method called MGDA, which leverages a learned environment dynamics model grounded in the local Lipschitz continuity assumption. We demonstrate that MGDA not only adheres to the established principles but also provides theoretical guarantees for the accurate prediction of augmented goal labels.
	\item In experiments conducted on offline maze datasets specifically designed to assess stitching capabilities, the integration of MGDA into GCWSL has demonstrated superior performance. Compared to other goal data augmentation methods such as Swapped Goal Data Augmentation (SGDA) \citep{yang2023swapped} and Temproal Goal Data Augmentation (TGDA) \citep{ghugare2024closing}, MGDA shows significant improvements when they are all added to GCWSL. Our further analysis and ablation studies highlight the crucial role played by the local Lipschitz continuity assumption in achieving these results.
\end{itemize}

\section{Related Work}
\textbf{The Stitching Property.}
The concept of stitching, as discussed by \citep{ziebart2008maximum}, is a characteristic property of temporal-difference (TD) learning algorithms such as those described by \citep{mnih2013playing,lillicrap2015continuous,fujimoto2018addressing,kostrikov2021offline}, which employ dynamic programming techniques. This property allows these algorithms to integrate data from various trajectories, thereby enhancing their effectiveness in managing complex tasks by leveraging historical data \citep{cheikhi2023statistical}.
On the other hand,
most SL-based RL methods lack this property.
\citep{ghugare2024closing} indicates from the perspective of combinatorial generalization that typical SL methods  
such as 
DT
\citep{chen2021decision}
and RvS
\citep{emmons2021rvs}
do not perform stitching.
The same situation also exists in offline GCRL
\citep{yang2023swapped}.
In offline GCRL,
GCWSL methods \citep{yang2022rethinking,ma2022far,hejna2023distance,sikchi2024score} find a single optimal trajectory corresponding to a given (state, goal) pair and have demonstrated strong performance in various goal-reaching tasks. However,
these methods are sub-optimal for some unseen skills and lack the ability to stitch information from multiple trajectories like most SL-based RL methods.
\\
\noindent
\textbf{Local Lipschitz Continuity in Environment Dynamics.} 
Local continuity ensures that small changes in actions or states result in correspondingly small changes in transitions, a characteristic that aligns with physical laws and is observed in the dynamics of many robotic systems and real-world scene.
And it in dynamics is commonly applied in classical control methods to ensure the existence and uniqueness of solutions to differential equations
\citep{li2004iterative,bonnard2007second}. This assumption is particularly valuable in nonlinear systems and is widely employed in robotic applications
\citep{seto1994adaptive, kahveci2007robust, sarangapani2018neural}. 
However, these methods often require pre-specified models and cost functions within an optimal control framework when leveraging dynamics continuity.
In contrast, we apply the local Lipschitz continuity assumption when learning the dynamics model to predict augmented goals for GCWSL agents. Perhaps the most comparable work to ours is that of \citep{ke2024ccil}, but a key difference is that we use this assumption solely for prediction rather than for generation, which may lead to greater estimation errors.
\\
\noindent
\textbf{Data Augmentation in RL.}
Data augmentation, recognized as an effective technique for enhancing generalization, has been widely applied in both RL \citep{srinivas2020curl, lu2020sampleefficient, stone2021distracting, kalashnikov2021mtopt, hansen2021generalisation, kostrikov2021image, yarats2021mastering} and SL \citep{Shorten2019aso}. We have observed that some methods \citep{char2022bats,yamagata2023qlearning, paster2023return} leverage dynamic programming to augment existing trajectories, thereby improving the performance of SL algorithms. However, these methods still rely on dynamic programming.
Another closely related approach focuses on data augmentation exclusively for SL \citep{yang2023swapped, ghugare2024closing} without employing dynamic programming. These methods are simple and efficient enough to enhance the stitching ability of the SL method. Nevertheless, our analysis indicates that they still encounter challenges in accurately providing augmented goal data.
\section{Background}
\textbf{Goal-conditioned Reinforcement Learning (GCRL)}.  
GCRL can be characterized by the tuple $\left\langle {\mathcal{S},\mathcal{A},\mathcal{G},\gamma,\rho_{0},T,f,r} \right\rangle$,
where 
$\mathcal{S}$,
$\mathcal{A}$,
$\mathcal{G}$,
$\gamma$,
$\rho_{0}$,
$T$
refer to state space,
action space,
goal space,
discounted factor,
the distribution of initial states and the horizon of the episode,
respectively.
$f:\mathcal{S} \times \mathcal{A} \to \mathcal{S}^{\prime}$
is the ground truth dynamic transition,
and $r:\mathcal{S} \times \mathcal{G} \times \mathcal{A} \to \mathbb{R}$ is typically a simple unshaped binary signal.
And the objective of $\pi(a|s,g)$ is maximizing returns of reaching goals from the goal distribution $p(g)$:
\begin{equation} \label{eq:2}
\mathcal{J}(\pi)=\mathbb{E}_{\substack{g\sim p(g), s_0\sim \rho_{0},\\ a_t\sim\pi,s_{t+1}\sim f(\cdot|s_t,a_t)}}\left[\sum_{t=0}^{\infty}\gamma^tr(s_t,a_t,g)\right].
\end{equation}

In this paper, the sparse reward function $r$ is defined as indicator function:
\begin{equation}\label{eq:1}
r(s_t,a_t,g)=\begin{cases}1,&\|\phi(s_t)-g\|<\delta\\0,&\text{otherwise}\end{cases},
\end{equation}
where $\delta$ is a threshold and
$\phi:\mathcal{S}\rightarrow \mathcal{G}$
is a known state-to-goal mapping function from states to goals.
And the goal is considered reached when $\|\phi(s_t)-g\|<\delta$
\citep{andrychowicz2017hindsight}. 

\textbf{Offline GCRL}. In offline RL setting,
the agent can only access a static offline dataset $\mathcal{D}$
and cannot interact with the environment to maximize the objective in Equation \ref{eq:2}.
The offline dataset $\mathcal{D}$ can be collected by some unknown policies
\citep{levine2020offline,prudencio2023survey}.
Based on the definition of GCRL,
we further express the offline dataset as ${\mathcal D}:=\{\tau_{i}\}_{i=1}^{N}$,
where $\tau_{i} :=\{<s_{0}^{i},a_{0}^{i},r_{0}^{i}>,<s_{1}^{i},a_{1}^{i},r_{1}^{i}>,...,<s_{T}^{i},a_{T}^{i},r_{T}^{i}>,g_{i}\}$ is the goal-conditioned trajectory and $N$ is the number of stored trajectories.
In $\tau_{i}$, 
$s_0 \sim \rho_0$.
The desired goal $g_{i}$ is still randomly sampled from $p(g)$.
Relabeled goals can be derived from each state as $g_t^i = \phi(s_t^i)$ for $ 0\leq t \leq T$. It should be noted that trajectories may be unsuccessful trajectories (i.e,
$g^i_T \neq g_{i}$).\\
\textbf{Goal-conditioned Weighted Supervised Learning (GCWSL)}. Different with general goal-conditioned RL methods that directly maximizes discounted cumulative return,
GCWSL provides theoretical guarantees that weighted supervised learning from
hindsight relabeled data optimizes a lower bound on the goal-conditioned RL objective in offline GCRL.
During training,
trajectories are sampled form a relabeled dataset by utilizing
hindsight mechanisms 
\citep{kaelbling1993learning,andrychowicz2017hindsight}.
And the policy optimization satisfies the following definition:
\begin{equation} \label{eq:3}
\mathcal{J}_{GCWSL}(\pi) = \mathbb{E}_{(s_{t},a_{t},g)\sim \mathcal{D}_{r}}\left[w\cdot\log\pi_{\theta}(a_{t}|s_{t},g)\right],
\end{equation}
where $\mathcal{D}_{r}$ denotes relabeled data,
$g=\phi(s_i)$ denotes the relabeled goal for
$i\geq t$.
The weighted function $w$ exists various forms in GCWSL methods
\citep{yang2022rethinking,ma2022offline,hejna2023distance,sikchi2024score} and can be
considered as the scheme choosing optimal path between $s$ and $g$.
Therefore GCWSL includes typical two process,
acquiring sub-trajectories corresponding to $(s,g)$ pairs and imitating them.
In the process of imitation,
GCWSL first train the specific weighted function $w$,
and then extract the policy with the Equation \ref{eq:3}.
\begin{figure}[t]
    \centering
    \begin{minipage}{\linewidth}
		\centerline{\includegraphics[width=7cm]{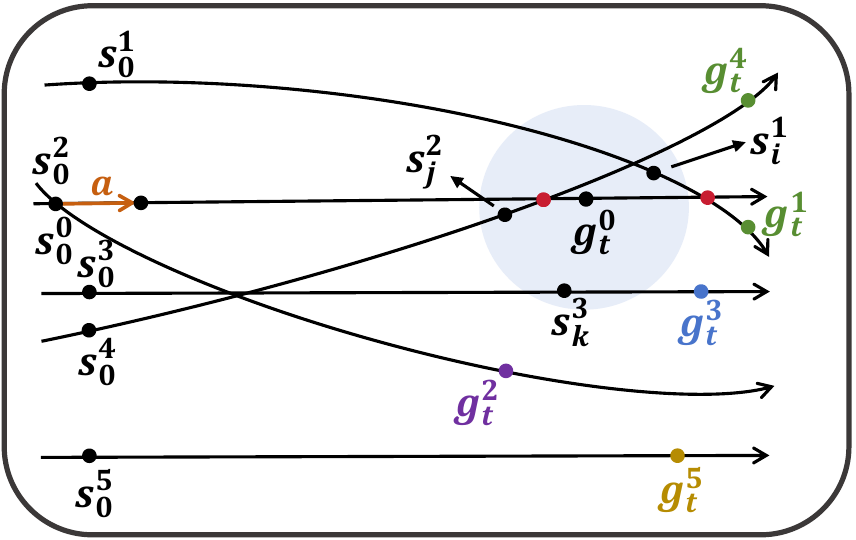}}
	\end{minipage}
    \begin{minipage}{0.7\linewidth}
		\centerline{\includegraphics[width=4.5cm]{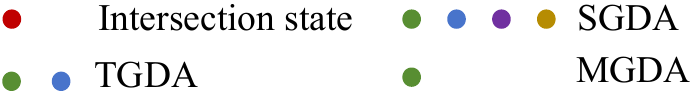}}
	\end{minipage}
    \caption{
    Counter examples of generalized principles and related goal data augmentation methods.
    The states $s^0_0$ through $s^5_0$ correspond to trajectories $\tau_0$ through $\tau_5$, respectively. 
    The goals $g^{0}_{t}$ through $g^{5}_{t}$ represent the relabeled goals for each respective trajectory.
    Note that $s^{0}_0$ is equal to $s^{2}_0$ and red points denote intersection state of two trajectories.
    The light blue circles represent the nearby states of $g^{0}_{t}$ after k-means clustering within the TGDA \citep{ghugare2024closing} .
    Compared to the original goal $g^{0}_{t}$, SGDA \citep{yang2023swapped} randomly select all goals 
    $g\in \left[g^{1}_{t},g^{2}_{t},g^{3}_{t},g^{4}_{t}, g^{5}_{t} \right]$
    as augmented goals, while TGDA selects the goals 
    $g\in \left[g^{1}_{t},g^{3}_{t},g^{4}_{t}\right]$
    from later in the trajectory corresponding to the nearby states 
    $s\in\left[s^{1}_{i},s^{2}_{j},s^{3}_{k}\right]$.
    Our MGDA method will select more appropriate goals 
    $g\in \left[g^{1}_{t},g^{4}_{t}\right]$, building upon TGDA by avoiding unreachable goals.
    }
	\label{fig:GCWSL}
\end{figure}
\section{Generalized Principles and Related Works of Goal Data Augmentation for GCWSL} \label{sc:4}
Data augmentation has become a common approach for improving generalization. In offline GCRL scenarios, it can also serve as an effective strategy to tackle unseen skills. Given the emergence of various methods, we utilize the characteristics of data augmentation and develop more general methods for competitive GCWSL methods in offline GCRL. This section proposes for the first time three principles of satisfying goal data augmentation for GCWSL methods. These principles also can be applied to other goal-conditioned SL methods. 
Moreover, 
we compare the differences among following goal augmentation methods with our MGDA using a counter example:
SGDA
\citep{yang2023swapped}
proposes a method that randomly choose augmented goals from different trajectories.
TGDA \citep{ghugare2024closing} proposed a another goal data augmentation approach from the perspective of combinatorial optimization. It employs k-means to cluster the goal and certain states into a group, and samples goals from later stages of these state trajectories as augmented goals.

Here we assume that we need to perform goal data augmentation for $(s^{0}_{0}, g^{0}_{t})$ pair as shown in Figure \ref{fig:GCWSL}. To select more appropriate augmentation goals for $g^{0}_{t}$, we established three guiding principles for goal data augmentation. Below, we elaborate on the specifics and motivation behind these three principles.\\
\textbf{\textit{Goal Diversity.}} This is a fundamental principle for goal data augmentation, which implies that for a given state $s^{0}_{0}$, multiple goals can be associated with it. These multiple goals should come from different trajectories.\\
\textbf{\textit{Action Optimality.}} This principle asserts that after selecting a goal, the original action corresponding to state $s^{0}_{0}$ can still be considered the optimal action for the augmented goal. 
It is also recognized in the TGDA \citep{ghugare2024closing}.
This principle helps to reduce the redundancy in goal data augmentation. In other words, the stitched trajectories of augmented goals can be treated as the optimal trajectories corresponding to the 
$(s^{0}_{0}, g^{0}_{t})$ pair, thereby simplifying the training complexity of GCWSL. For instance, we choose $g^{2}_{t}$ as the augmented goal for the original state $s^{0}_{0}$. Even if it is reachable for $s^{0}_{0}$, the optimal action from 
$s^{0}_{0}$ to $g^{2}_{t}$ no longer align with the original action 
$a$. In this way $g^{2}_{t}$ cannot be used as an augmented goal for $g^{0}_{t}$, but only as another goal for $s^{2}_{0}$ that needs to be reached by re-learning. Since SGDA selected $g^{2}_{t}$, it does not satisfy \textit{Action Optimality.}\\
\textbf{\textit{Goal Reachability.}} This principle emphasizes that when selecting augmented goals from other trajectories for the original state 
$s^{0}_{0}$, the chosen goals must be reachable. Specifically, the original trajectory and other trajectories must share a intersection state. For instance, in Figure \ref{fig:GCWSL}, trajectories $\tau_3$ and $\tau_5$ do not intersect with the original trajectory $\tau_0$, rendering the chosen goals 
$\left[g^{3}_{t},g^{5}_{t}\right]$
ineffective.
SGDA exists cases where the augmented goal data are unreachable such
as $\left[g^3_t,g^5_t\right]$.
TGDA also faces the existence of unreachable goals. 
As shown in Figure \ref{fig:GCWSL}, assume a wall between trajectory $\tau_0$ and $\tau_3$, k-means might still identify unreachable goals $g^3_t$ as augmented goals, which is unreasonable. 

In summary, \textbf{\textit{Goal Diversity}} is a fundamental guideline that most data augmentation methods aim to achieve, while \textbf{\textit{Action Optimality}} and \textbf{\textit{Goal Reachability}} are specifically tailored for offline GCRL, building upon \textbf{\textit{Goal Diversity}}. 
Our MGDA approach overcomes the limitations of previous goal data augmentation methods and adheres to all three principles.
Detailed comparisons can be found in Table \ref{augment table}. 
\begin{table}[h]
   \centering
    \renewcommand\arraystretch{1.3}
    \tabcolsep=0.05cm
    \scalebox{0.8}{\begin{tabular}{ll|lll}
    \toprule
    \multicolumn{1}{c}{\textbf{Methods}} & \multicolumn{1}{c|}{\textbf{Metric}} & \makecell[c]{\textbf{\textit{Goal Diversity}}} & \makecell[c]{\textbf{\textit{Action Optimality}}} & \makecell[c]{\textbf{\textit{Goal Reachability}}}  \\
    \midrule
    \multicolumn{1}{c}{SGDA} & \makecell[c]{-} & \makecell[c]{\Checkmark}             &\makecell[c]{\XSolidBrush}               & \makecell[c]{\XSolidBrush}   \\
    \multicolumn{1}{c}{TGDA} & \makecell[c]{L2} & \makecell[c]{\Checkmark}                & \makecell[c]{\Checkmark}               & \makecell[c]{\XSolidBrush}   \\
    \multicolumn{1}{c}{\textbf{MGDA}} & \makecell[c]{Dynamics} & \makecell[c]{\Checkmark}              & \makecell[c]{\Checkmark}               & \makecell[c]{\Checkmark}    \\
    \bottomrule
    \end{tabular}}%
    \caption{Comparison of different goal data augmentation methods for GCWSL.}
    \label{augment table}%
\end{table}

\section{MGDA: Model-based Goal Data Augmentation}
In this section, we describe MGDA in detail. The core idea is to \textbf{identify nearby states relative to the original goal and then sample an augmented goal from the latter part of the trajectory corresponding to these nearby states to serve an augmentation of the original goal. And these nearby states must successfully reach the goal}. In this case, the original goal functions as an intersection state, linking two different trajectories.

We begin by detailing the process of learning a dynamics model that leverages the local Lipschitz continuity assumption to accurately predict nearby states and sample augmented goals. Following this, we provide both practical implementation and theoretical justification for MGDA.
\subsection{Sample Augmented Goals with MGDA}
A critical step in our approach involves learning the dynamics model from data and utilizing it for state prediction by leveraging the local continuity of the environment. When the dynamics model is locally Lipschitz bounded, small variations in state and action lead to correspondingly small changes in transitions. A dynamics function with local continuity enables us to accurately identify nearby states around the original goal that satisfy the dynamic transition conditions observed in the training data within this region, the reliability of the learned model can be assured.
\begin{figure}[!h]
   \centering
   \begin{minipage}{\linewidth}
		\centerline{\includegraphics[width=5cm]{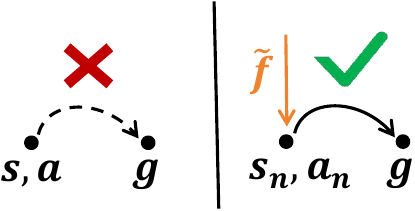}}
	\end{minipage}
   \begin{minipage}{\linewidth}
        \hspace{0.4cm}
		\centerline{\includegraphics[width=6.0cm]{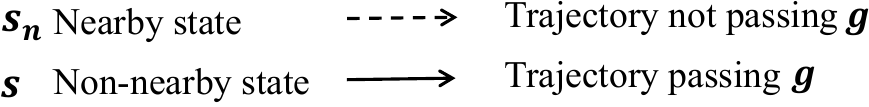}}
    \end{minipage}
   \caption{\footnotesize
   (\emph{left}) State searched by other goal data augmentation methods.
   (\emph{right}) State searched by our MGDA, constrained by the dynamics model in its relationship to the goal.
   MGDA ensures that the searched state and goal $g$ satisfy the one-step transition criterion, thereby defining this searched state as a nearby state $s_n$.
   }
   \label{fig:augment goal}
\end{figure}
We identify the nearby state under the metric defined by the dynamics model, as illustrated in Figure \ref{fig:augment goal}. In this figure, we denote the correct nearby state as $s_n$ and the learned dynamics function is represented as $\hat{f}$. Additionally, the action corresponding to $s_n$ is denoted as $a_n$. To guarantee accurate predictions of nearby state $s_n$,
we follow the model-based reinforcement learning frameworks proposed in \citet{gulrajani2017improved} and \citet{ke2024ccil} to enforce local continuity on the dynamics model's learning process:
\begin{align} \label{eq:5}
\arg\min_{\hat{f}}\mathbb{E}_{s_n,a_n,g\sim\mathcal{D}}\bigl[\sigma(\lambda_{n})\cdot\|g-s_{n}-\hat{f}(s_{n},a_{n})\| \\ \nonumber
+\sum\sigma(\lambda_{n})\bigr], \mathrm{while}~W\to W/\max(\frac{||W||_{2}}{\lambda},1)
\end{align}
where $\lambda_n$ is a state-dependent slack variable used to dynamically adjust the weight of each data point in the loss function, $\lambda$ is a global regularization parameter in the training objective that controls the overall penalty strength for enforcing local Lipschitz continuity in the model,
$\sigma$ is the Sigmod function and $W$ is a weight matrix that scales the dynamics model's predictions. 
Equation \ref{eq:5} introduces a modified form of the mean squared error (MSE) loss, which we employ to fit the function $\hat{f}$ with a neural network and predict the changes in both the nearby state $s_n$ and the goal $g$.

Then we can ensure that the approximate model remains predominantly  $L$- Lipschitz bounded while accurately predicting transitions in the given offline data. As a result, this approximate dynamics model can be effectively used to predict corrective labels.
\begin{theorem}[model smoothness]\label{theorem:1} 
As described in Figure \ref{fig:augment goal}, under the assumption of local Lipschitz continuity, when the error generated by training the model is $\epsilon$, the 1-step residual dynamics model $\hat{f}$ is subject to the following boundary for predicting the correct nearby states $s_n$ of goal $g$.
\begin{equation} 
\left\|f\left(s_n,a_n\right)-\hat{f}\left(s_n,a_n\right)\right\|\leq\epsilon+\big(K+\Delta(\lambda_n)\big)\left\|s_n-g\right\|,
\end{equation}
where $K$ is the Lipschitz constants for true environment dynamics $f$ and $\Delta(\lambda_n)$ is the Lipschitz error controlled by $\lambda_n$ for learned dynamics model $\hat{f}$.
\end{theorem}
The full proof can be found at supplementary material. This theorem demonstrates that when the dynamics function is governed by a local Lipschitz constant, the prediction error of the dynamics model can be quantified accordingly based on that constant.

After searching for the nearby states of original goal, we sample a new augmented goal from later in that trajectory, as shown in Figure \ref{fig:GCWSL}.
This approach provides a straightforward method for sampling cross trajectory goals while ensuring that the action remains optimal at original state.

\subsection{Practical Implementation of MGDA and Analysis}
To expedite the identification of nearby states, our MGDA algorithm first employs the k-means technique to cluster all states into multiple groups, as proposed by \citep{ghugare2024closing}. Subsequently, we train a transition model on the data and search for nearby states within the permissible range of model error.
Specifically, we first randomly select a initial state within the same group as the original goal and apply the dynamic function for joint forward inference to generate the resulting state. And then if this resulting state reach the original goal
(i.e, the L2 distance between the resulting state and the original goal is less than $\delta$)
, we designate this initial state as nearby state to original goal.
Finally, a new goal is randomly chosen as an augmentation goal for original goal from the trajectory following nearby state.
Therefore MGDA can be considered as model-augmented version of TGDA.
The complete algorithm for adding MGDA to GCWSL is illustrated in Algorithm \ref{alg:MGDA}. The blue text represents the entire process of MGDA.
\begin{algorithm}[h]
	\caption{MGDA for GCWSL Methods}
	\label{alg:MGDA}
	\begin{algorithmic}[1]
		\STATE \textbf{Input:}
		Offline Dataset $\mathcal{D}$.
        \STATE \textbf{Initialize:} Policy $\pi_{\theta}(a|s,g)$ with parameter $\theta$.
        {\color{blue}
         \STATE \textbf{Function} Learn Dynamics $\hat{f}$
         \STATE \quad Optimize the objective in Equation \ref{eq:5}.
        }
		\WHILE{a fixed number of iteration}
            \FOR{$t=1,\cdots,m$}
                \STATE Relabel and sample $\{(s_t,a_t,g)\sim \mathcal{D}$,
                where $g=\phi(s_{i}), i\geq t$.
                {\color{blue}
                \STATE \indent Get the group of the goal with k-means: $k = d_{t+}$,
                where $d_l$ = CLUSTER($s_l$).
                \STATE \indent Randomly sample candidate state from the same group:  $u \sim \{s_j; \forall j \; \text{such that}\; d_j = k \}$.
                \IF {$\|g-u-\hat{f}(s_t,a_t)\|\textless \delta$}
                       \STATE Identify $s_n := u$ as a nearby state.
                       \STATE Sample new goal $\tilde{g}$ from the later stages in the trajectory of $s_n$.
                       \STATE Augment the goal $g = \tilde{g}$.
                \ENDIF
                }
            \STATE Calculate the weight $w$ in original GCWSL methods and Update $\pi_{\theta}$ with maximize goal-conditioned policy  optimization with Equation \ref{eq:3}.
        \ENDFOR
        \ENDWHILE
        \STATE \textbf{Return:} 
        Goal-conditioned policy $\pi_{\theta}(a|s,g)$
	\end{algorithmic}
 \label{alg:MGDA}
\end{algorithm}

While data augmentation methods generally lack theoretical guarantees, we demonstrate that MGDA approximates a one-step stitching policy under the smooth assumption of certain distributions. The proposed new model-based method MGDA can generate (state, goal) combinations that were not encountered during training, effectively mimicking the stitching process observed in TD learning.

If we assume that our offline dataset ${\mathcal D}:=\{s^i_0,a^i_0,...\}_{i=1}^{N}$ is collected by a set of policies $\{\beta(a\mid s,h)\}$ where $h$ specifies some context from distribution $p(h)$ and $\beta_h:=\beta_h(a|s,h)$ \citep{ghugare2024closing}.
We denote $p_{+}^{\pi}(g\mid s,a)\triangleq(1-\gamma)\sum_{t=0}^{\infty}\gamma^{t}p_{t}^{\pi}(s_{t}=g\mid s_{0}=s)$ is the discounted state occupancy distribution for a goal-conditioned policy $\pi$. We first have follow assumption (Drawing inspiration from Appendix D.2 in \citet{ghugare2024closing}):
\begin{assumption}[distribution smoothness]
For all $s, a, g$ pairs and all data collecting policies $\beta_h$,
$s_n$ and $s^{\prime}_n$ are the states corresponding to the reachable and unreachable goals in $u$, respectively.
The $p_{+}^{\beta_{h}}(g\mid s,a)$ is $L_1$-Lipschitz continuous with respect to 
$s_n$ and $L_2$-Lipschitz continuous with respect to $s^{\prime}_n$:
\begin{equation} \label{eq:7}
|p_+^{\beta_h}(g\mid s,a)-p_+^{\beta_h}(s_n\mid s,a)|\leq L_1(||g-s_n||)
\end{equation}
and:
\begin{equation} \label{eq:8}
|p_+^{\beta_h}(g\mid s,a)-p_+^{\beta_h}(s^{\prime}_n\mid s,a)|\leq L_2(||g-s^{\prime}_n||)
\end{equation}
\end{assumption}
Intuitively, the above all smoothness result ensures that all intra-group states under the data collection policy have similar probabilities. For example, after taking action $a$ from state $s$, there is an equal probability of reaching nearby states $s_n$ and $g$.
And then we have follow guarantee for MGDA:
\begin{theorem} \label{theorem:2}
Given the aforementioned smoothness assumption, the model-based goal data augmentation $p^\text{MGDA}(g\mid s,a)$ approximates the sampling process of goals according to the distribution defined by the one-step goal-reaching stitching policy $p^\text{1-step}(g\mid s,a)$ for all $s, a, g$ pairs:
\begin{equation}
p^\text{MGDA}(g\mid s,a) = p^\text{1-step}(g\mid s,a)\pm\mathcal{O}(\epsilon_k L_1),
\end{equation}
where $\epsilon_k$ is the maximum cutoff distance between all states within the group of $g$ after clustering.
\end{theorem}
This theorem demonstrates that a single application of model-based goal data augmentation samples $(s,g)$ pairs from the 1-step goal-reaching distribution.
\section{Experiments}
\begin{figure*}[!h]
    \centering
	\begin{minipage}{\linewidth}
		\centerline{\includegraphics[width=8.5cm]{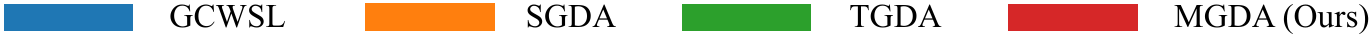}}
    \end{minipage}
    \begin{minipage}{0.32\linewidth}
		\centerline{\includegraphics[width=\textwidth]{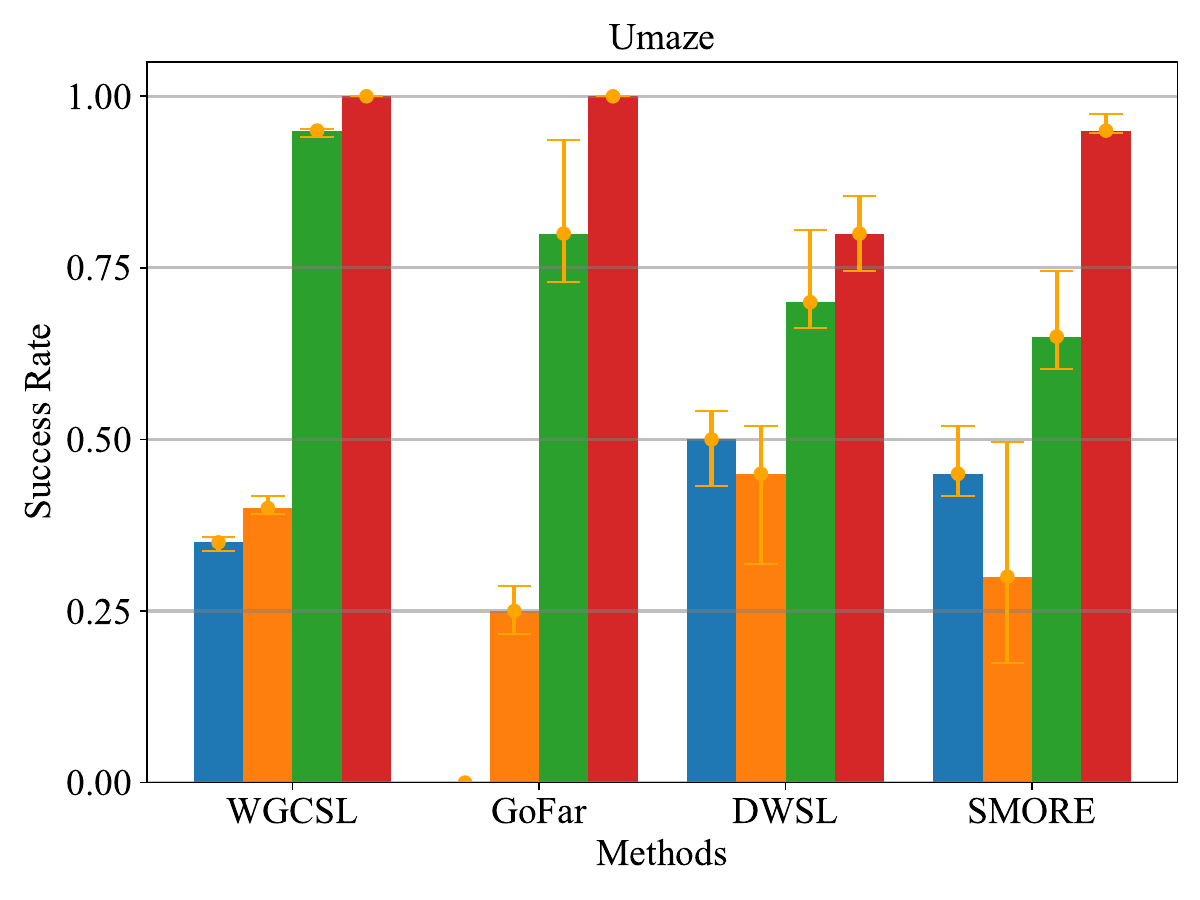}}
	\end{minipage}
    \begin{minipage}{0.32\linewidth}
		\centerline{\includegraphics[width=0.95\textwidth]{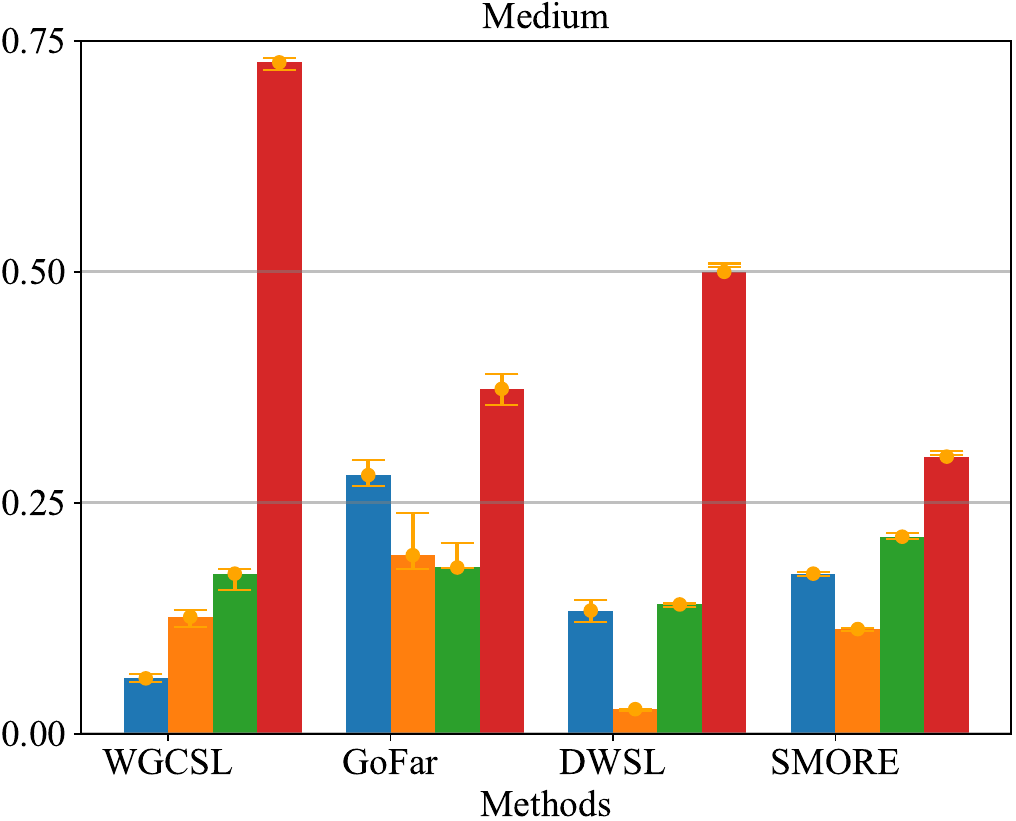}}
	\end{minipage}
	\begin{minipage}{0.32\linewidth}
		\centerline{\includegraphics[width=0.95\textwidth]{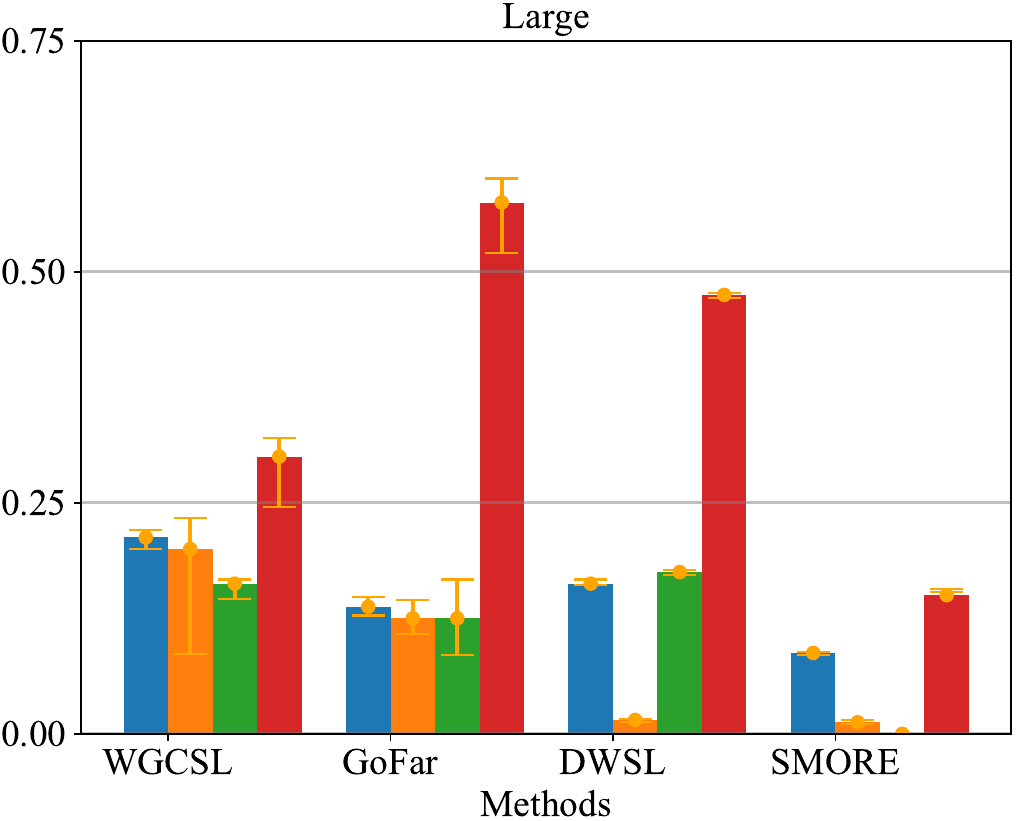}}
	\end{minipage}
    \caption{
     Performance of the original GCWSL methods and the impact of incorporating different goal augmentation approaches in state-based datasets. We use the final mean success rate as the report.  Error bars denote 95$\%$ bootstrap confidence intervals.
    }
    \label{fig:state goal results}
\end{figure*}
\begin{figure*}[!h]
    \centering
    \begin{minipage}{\linewidth}
		\centerline{\includegraphics[width=8.5cm]{results_pdf/results1.pdf}}
    \end{minipage}
	\begin{minipage}{0.32\linewidth}
		\centerline{\includegraphics[width=\textwidth]{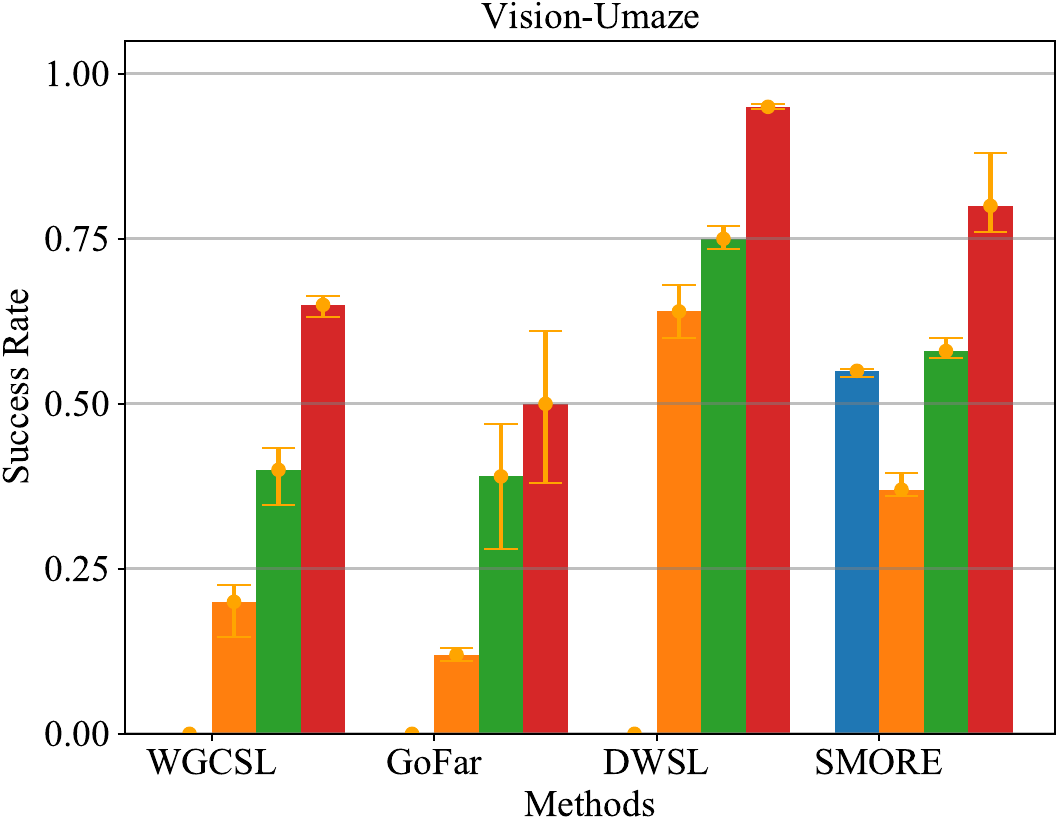}}
	\end{minipage}
    \begin{minipage}{0.32\linewidth}
		\centerline{\includegraphics[width=0.95\textwidth]{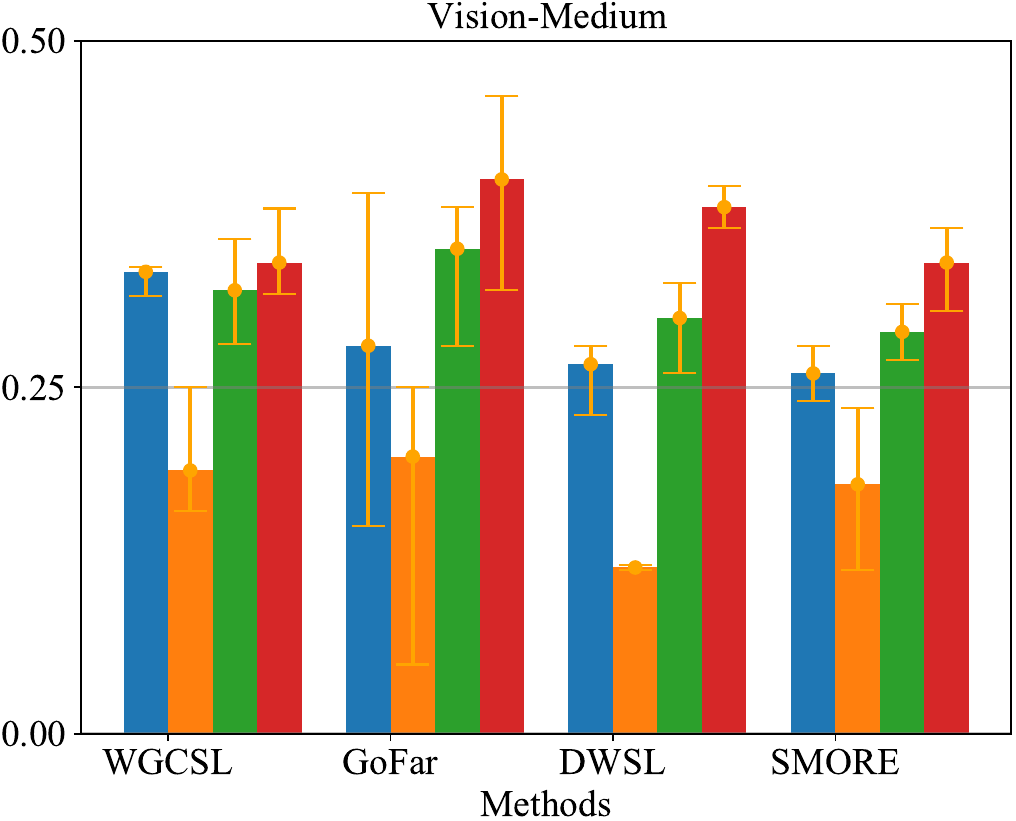}}
	\end{minipage}
	\begin{minipage}{0.32\linewidth}
		\centerline{\includegraphics[width=0.95\textwidth]{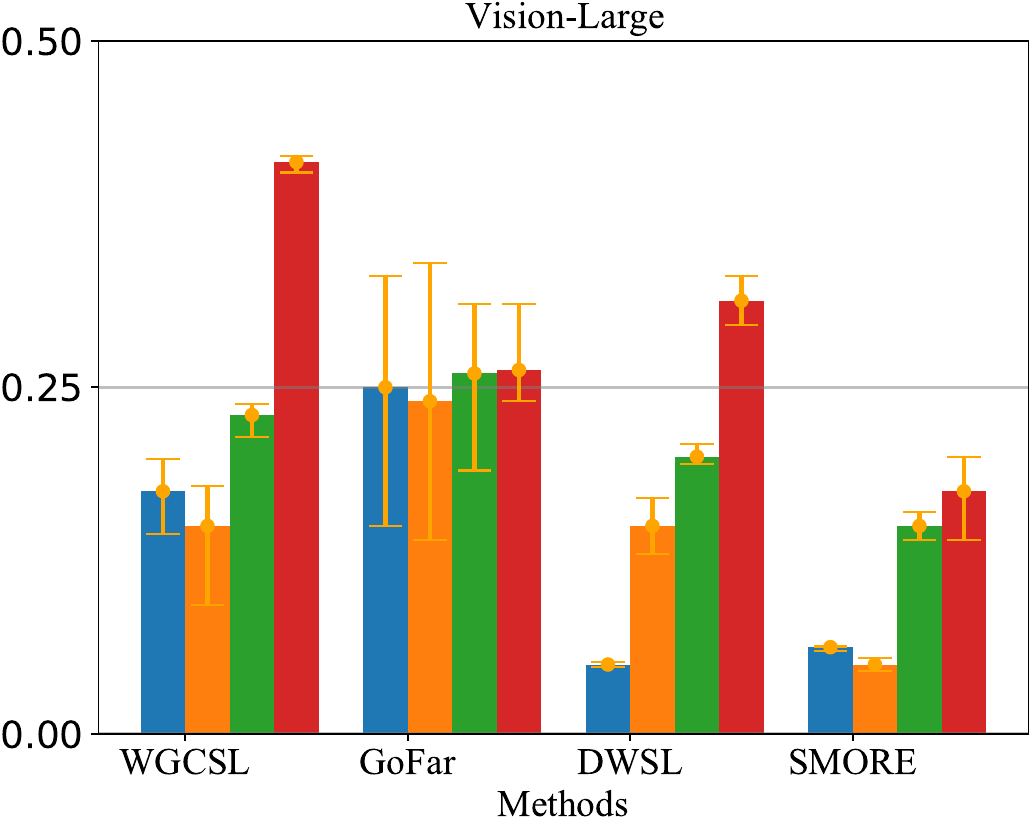}}
	\end{minipage}
    \caption{
     Performance comparison between the original GCWSL approach and its enhancement with MGDA on vision-based datasets. We also use the final mean success rate as the report. Error bars denote 95$\%$ bootstrap confidence intervals.
    }
    \label{fig:vision goal results}
\end{figure*}
\textbf{Datasets}. To rigorously evaluate the stitching capabilities of GCWSL methods, we employ the offline point maze dataset configuration as outlined in \citep{ghugare2024closing}. For this evaluation, we modify the GCWSL policy to navigate between previously unseen combinatorial (state, goal) pairs and subsequently measure the success rate.\\
\textbf{Baselines}.
We conducted a series of comparative experiments by implementing the GCWSL methods within the same framework,
as well as related goal data augmentation approaches. Specifically, all GCWSL implementations are based on DWSL \citep{hejna2023distance}, with hyperparameter values set to the default values specified in the original studies. We consider four competitive baseline GCWSL algorithms: \textbf{WGCSL} \citep{yang2022rethinking}, \textbf{GoFar} \citep{ma2022offline}, \textbf{DWSL} \citep{hejna2023distance}, and \textbf{SMORE} \citep{sikchi2024score}. 

All goal data augmentation implementations including \textbf{SGDA} \citep{yang2023swapped} and \textbf{TGDA} \citep{ghugare2024closing} are based on TGDA \citep{ghugare2024closing}, 
while utilizing the same augmentation probability.
All experiments are conducted using five random seeds. Detailed algorithm implementations and hyperparameter settings are provided in the supplementary material.
\section{Experimental Results}
In this section we seek to answer the following questions: 1) \emph{How does GCWSL perform when combined with our MGDA method? Does more data remove the need for goal data augmentation?} 2) \emph{Can MGDA be effective for high-dimensional tasks? } 3) \emph{How does MGDA compare in performance to other data augmentation methods?} 4) \emph{How essential is the local Lipschitz continuity assumption?}
\\
\textbf{State-based Dataset Results}. As shown in Figure \ref{fig:state goal results}, it is evident that all GCWSL algorithms struggle to demonstrate stitching properties, particularly in the complex Medium and Large tasks, where their performance is notably poor. However, when MGDA is incorporated into the GCWSL methods, performance improvements were observed across all tasks, albeit to varying degrees. This enhancement is attributed to the fact that goal data augmentation allows for the sampling of unseen (state, goal) combinations during the training phase, thereby improving the generalization and stitching capabilities of the models.
\\
\textbf{Vision-based Datasets Results.} Figure \ref{fig:vision goal results} demonstrates that MGDA similarly enhances the performance of all GCWSL methods, indicating its effectiveness in high-dimensional goal-reaching tasks and its overall robustness. However, in the complex Vision-Medium and Vision-Large tasks, the results were inconsistent, with some instances showing minimal improvement over the original GCWSL methods. This suggests that while MGDA offers benefits, its robustness may be limited in certain scenarios, and future research should focus on developing more robust and scalable approaches to address these challenges.
\\
\textbf{Comparison with Related Goal Data Augmentation Approaches.}
\begin{figure*}[!h]
    \centering
    \begin{minipage}{\linewidth}
		\centerline{\includegraphics[width=6.5cm]{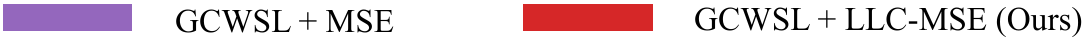}}
    \end{minipage}
	\begin{minipage}{0.32\linewidth}
		\centerline{\includegraphics[width=\textwidth]{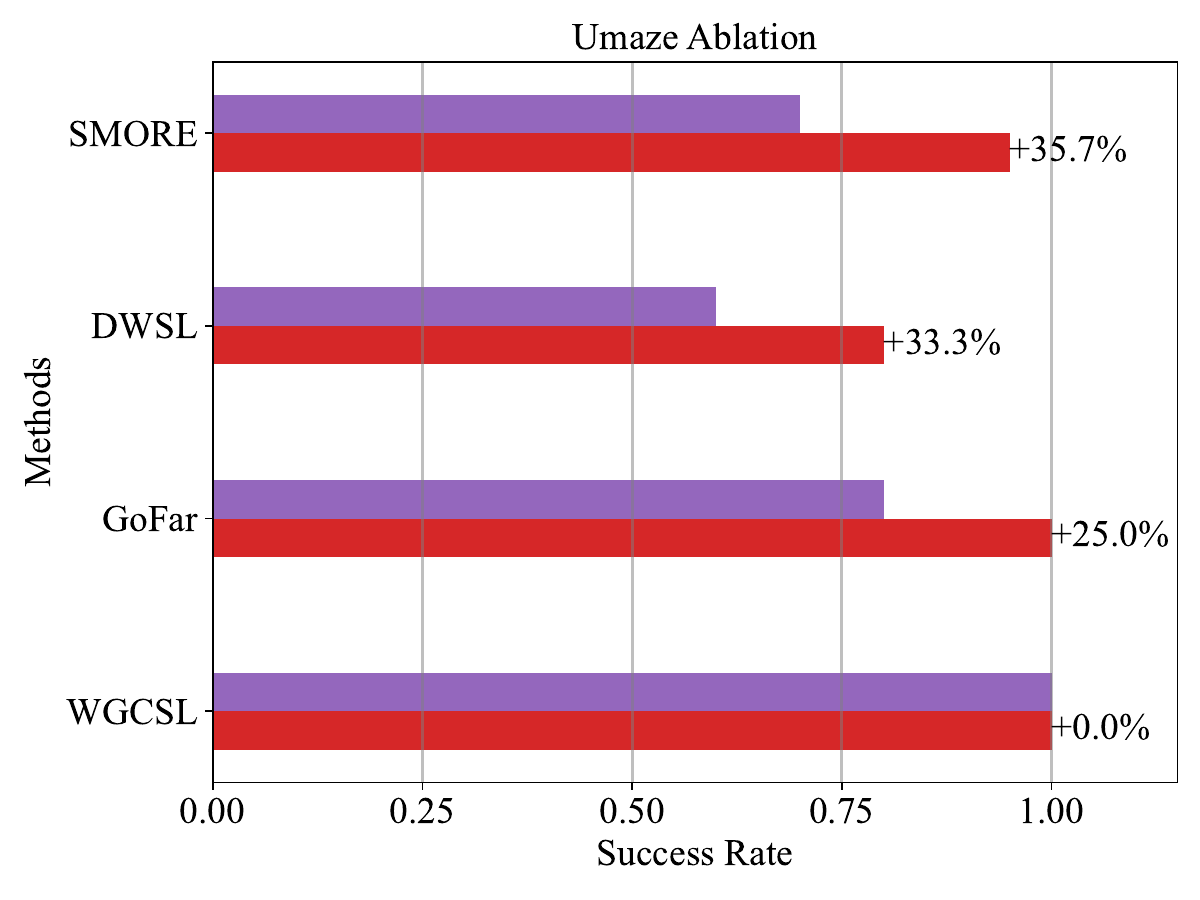}}
	\end{minipage}
    \begin{minipage}{0.32\linewidth}
		\centerline{\includegraphics[width=0.97\textwidth]{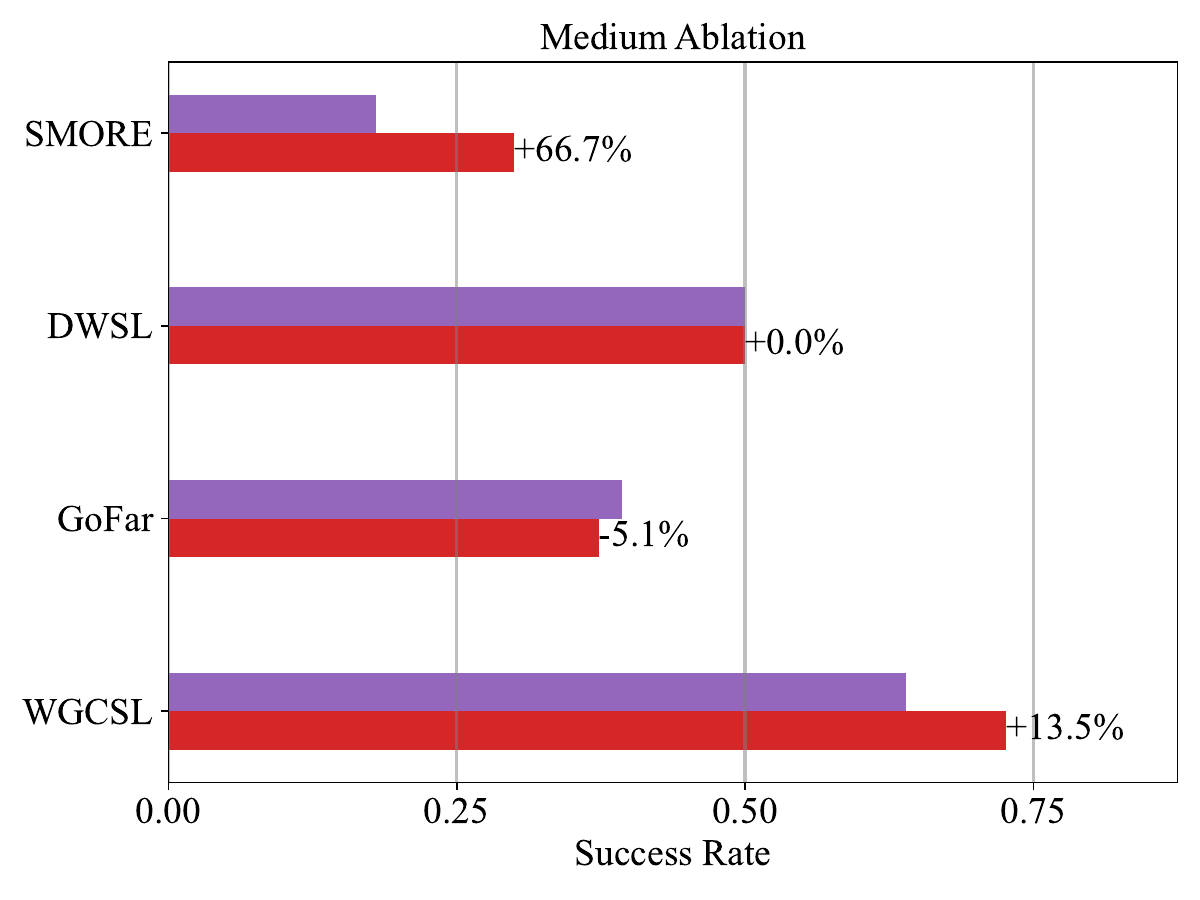}}
	\end{minipage}
	\begin{minipage}{0.32\linewidth}
		\centerline{\includegraphics[width=0.97\textwidth]{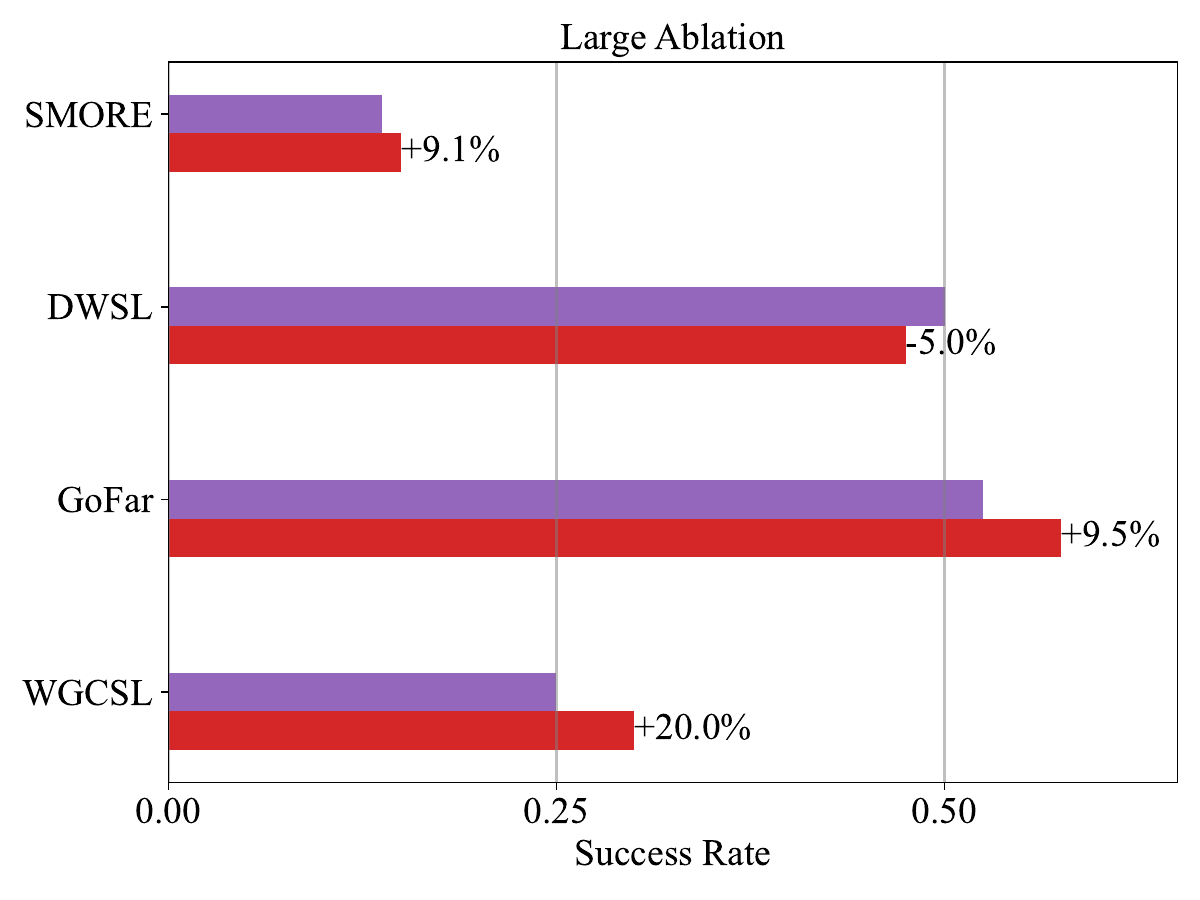}}
	\end{minipage}
    \caption{
     Ablation study on the local Lipschitz continuity (LLC) assumption. The results clearly show that the modified MSE version, incorporating LLC, outperforms the standard MSE-based dynamics model. This highlights the crucial role of the LLC assumption in enhancing performance.
    }
    \label{fig:importance goal results}
\end{figure*}
\begin{table*}[t]
    \vspace{-0.5em}
    \centering
    \scalebox{0.9}{
    \begin{tabular}{r|rrrr|rrrr}
    \toprule
    \multicolumn{1}{c|}{\multirow{2}[1]{*}{Dataset}} & \multicolumn{4}{c|}{DWSL with Different Dataset Size} &\multicolumn{1}{c}{\multirow{2}[4]{*}{\textbf{DWSL ($10^5$)}}} &\multicolumn{1}{c}{\multirow{2}[4]{*}{\textbf{DWSL ($10^6$)}}} &\multicolumn{1}{c}{\multirow{2}[4]{*}{\textbf{DWSL ($10^7$)}}} &\multicolumn{1}{c}{\multirow{2}[4]{*}{\textbf{DWSL ($10^8$)}}} \\
    \cmidrule{2-5}          & \multicolumn{1}{c}{$10^5$} & \multicolumn{1}{c}{$10^6$} & \multicolumn{1}{c}{$10^7$} & \multicolumn{1}{c|}{$10^8$}  \\
    \midrule
    \multicolumn{1}{l|}{Umaze} & 50.2$\pm$3.9  & 50.6$\pm$8.0 & 55.5$\pm$4.5 & 63.7 $\pm$2.9   & \colorbox{magiccolor}{72.4$\pm$3.3} & \colorbox{magiccolor}{80.7$\pm$2.7} & \colorbox{magiccolor}{87.6$\pm$8.6} & \colorbox{magiccolor}{91.2$\pm$1.8} \\
    \multicolumn{1}{l|}{Medium} & 21.3$\pm$2.3 & 22.3$\pm$2.3 & 24.1$\pm$1.5 & 28.3$\pm$5.8   & \colorbox{magiccolor}{52.6$\pm$9.2} & \colorbox{magiccolor}{56.8$\pm$5.3} & \colorbox{magiccolor}{58.2$\pm$6.1} & \colorbox{magiccolor}{63.4$\pm$2.6}  \\
    \multicolumn{1}{l|}{Large} & 11.3$\pm$4.9  & 16.3$\pm$6.7 & 15.8$\pm$8.9  & 21.3  $\pm$7.3    & \colorbox{magiccolor}{36.7$\pm$4.5} & \colorbox{magiccolor}{47.5$\pm$6.1} & \colorbox{magiccolor}{52.5$\pm$1.8} & \colorbox{magiccolor}{56.1$\pm$3.9}  \\
   \cmidrule{1-9}    \textit{Total} & \textit{82.8} & \textit{89.2} & \textit{95.4} &\textit{113.3}  & \colorbox{magiccolor}{161.7} & \colorbox{magiccolor}{\textit{179.0}} & \colorbox{magiccolor}{198.3} & \colorbox{magiccolor}{210.7}  \\
    \bottomrule
    \end{tabular}}%
    \caption{\footnotesize \label{tab2} We report the mean success rate ($\%$) across different dataset sizes, computed over five seeds, with each seed evaluated on 100 trajectories. \textbf{DWSL ($10^n, n=5,6,7,8$)} indicates the performance of the DWSL algorithm with MGDA integration on datasets of size $10^n$. }
\end{table*}
Based on previous discussion, we selected SGDA and TGDA as baseline methods, as outlined in Table \ref{augment table}. As shown in Figure \ref{fig:state goal results}, our MGDA method consistently outperforms the other data augmentation approaches across all tasks, particularly in the more complex Medium and Large datasets. 
SGDA consistently exhibits poor performance across all tasks. While TGDA performs well in the Umaze dataset, it fails to achieve comparable results in the more complex Medium and Large tasks, often underperforming relative to SGDA, which suggests a lack of robustness. In contrast, MGDA not only achieves superior performance across all tasks but also demonstrates greater robustness compared to both SGDA and TGDA.
\\
\textbf{Ablation Study.} Here we investigate the necessity of goal data augmentation for the highly-regarded algorithm DWSL \citep{hejna2023distance} within the GCWSL framework. The conventional wisdom suggests that larger datasets generally result in better generalization. To
empirically test whether this is the case, we trained original DWSL on four different dataset sizes and then adding it with MGDA on 10 million transitions. As shown in Table \ref{tab2}, increasing the dataset size did not result in improved performance for DWSL. Due to space constraints, additional results are provided in the supplementary material. Based on the results of all necessary experiments, simply increasing the size of the dataset does not enhance the generalization capability of the GCWSL methods. This suggests that traditional methods of expanding datasets may not effectively increase trajectory diversity. However, goal data augmentation can address this issue.

Finally, to assess the significance of the local Lipschitz continuity assumption, we compared MGDA with a variant of goal data augmentation that relies solely on standard MSE loss for learning the dynamics model, thereby omitting the local Lipschitz continuity assumption. As shown in Figure \ref{fig:importance goal results}, the local Lipschitz continuity assumption generally leads to superior performance across most goal-reaching tasks when compared to the MSE-based dynamics model learning method.
Although there are instances where the MSE approach outperforms our continuity-based model method in certain datasets and algorithms, we contend that incorporating local continuity in dynamics learning is valuable and merits further exploration. In summary, the local Lipschitz continuity assumption plays a crucial role in the effective learning of dynamics model.
\section{Conclusion and Future Work}
In this paper, we explored goal data augmentation methods for advanced GCWSL algorithms. While these methods enable GCWSL to exhibit stitching capabilities, existing approaches still struggle to select appropriate augmented goals for GCWSL. To address this issue, we proposed a unified set of goal data augmentation principles specifically tailored for GCWSL methods and introduced a novel model-based approach called MGDA. MGDA leverages the local Lipschitz continuity assumption to quantify model learning errors, facilitating more accurate prediction of augmented goals. Both theoretical analysis and experiments indicate the effectiveness of our proposed method MGDA.

However, this work has two primary limitations. First, our experiments indicate that the effectiveness of MGDA may be influenced by the number of layers in the neural network. Second, in certain offline datasets, MGDA did not consistently outperform existing data augmentation methods. Future research should explore more robust and scalable approaches to further enhance performance.

\section*{Acknowledgments}
This work was supported by the STI2030-Major Projects under Grant No. 2022ZD0208800. This research was supported by the Provincial Key Research and Development Program (Project No. 2023-YBGY-033).
\bibliography{aaai25}

\newpage
\onecolumn
\setcounter{section}{0}
\renewcommand\thesection{\Alph{section}} 

{\centering\section*{\LARGE\bf Supplementary Material}}
In this Supplementary Material, we provide more elaboration on the implementation details, results, and analysis.

\section{Theory Derivation}
\subsection{Proof of Theorem \ref{theorem:1}}
\label{sec:apx-thm1}
\textbf{Notation}.
Let $f$ be the ground truth 1-step residual dynamics model, and let $\hat{f}$ be the learned approximation of $f$.Given a goal $g$ in offline data,
we aim to predict the neighboring states $s_n$ around $g$.
Note that there is a known mapping relationship between goal space and state space:
$g=\phi(s), \phi:\mathcal{S}\rightarrow \mathcal{G}$.\\
\textbf{Assumptions}.
Before proving this theorem, we first have the following assumptions:
\\
\noindent
(1) The error between a dynamics model $\hat{f}$ learned on an offline dataset and the true model $f$ is defined by the following boundaries:
\begin{equation}
\left\|f(s_n,a)-\hat{f}(s_n,a)\right\|\leq\epsilon.
\end{equation}
\\
\noindent
(2) $f$ is locally $K$-Lipschitz continuous at nearby state $s_n$ around $g$ and satisfies the following bound:
\begin{equation}
\left\|f(s_n,a)-\hat{f}(g,a)\right\|\leq K\left\|s_n-g\right\|.
\end{equation}
\\
\noindent
(3) $\hat{f}$ is also locally $\Delta(\lambda_{n})$-Lipschitz continuous nearby state $s_n$ around $g$ and satisfies the following bound:
\begin{equation}
\left\|f(s_n,a)-\hat{f}(g,a)\right\|\leq \Delta(\lambda_{n}) \left\|s_n-g\right\|.
\end{equation}
\\
\noindent
And then we have the following proof.
\\
\noindent
\textbf{Proof}.\\ 
\begin{flalign}
&\
\left\|f(s_n,a)-\hat{f}(s_n,a)\right\|\nonumber\\
&=\left\|f(s_n,a)-f(g,a)+f(g,a)-\hat{f}(g,a)+\hat{f}(g,a)-\hat{f}(s_n,a)\right\|\nonumber\\
&\leq\left\|f(s_n,a)-f(g,a)\right\|+\left\|f(g,a)-\hat{f}(g,a)\right\|+\left\|\hat{f}(g,a)-\hat{f}(s_n,a)\right\|\nonumber\\
&\leq\epsilon+\big(K+\Delta(\lambda_{n})\big)\left\|s_n-g\right\|
&
\end{flalign}
\subsection{Proof of Theorem \ref{theorem:2}}
\textbf{Proof}( Rephrased from Lemma D.2 of \citep{ghugare2024closing}).
\begin{flalign}
&\
p^{\mathrm{MGDA}}(g\mid s,a) \nonumber\\
&=\int_{u}\sum_{h}p(h\mid s,a)(p_{+}^{\mathrm{MGDA}}(w\mid s,a))\sum_{\tilde{h}}p(\tilde{h}\mid u)p_{+}^{\beta_{\tilde{h}}}(g\mid w)du \nonumber\\
&\stackrel{a}{=}\int_{s_{n}}\sum_{h}p(h\mid s,a)(p_{+}^{\beta_{h}}(s_{n}\mid s,a)\pm\epsilon_k L_1)\sum_{\tilde{h}}p(\tilde{h}\mid s_{n})p_{+}^{\beta_{\bar{h}}}(g\mid s_{n})ds_{n} \nonumber\\
&+\cancel{\int_{s^{\prime}_{n}}\sum_{h}p(h\mid s,a)(p_{+}^{\beta_{h}}(s^{\prime}_{n}\mid s,a)\pm\epsilon_k L_2)\sum_{\tilde{h}}p(\tilde{h}\mid s^{\prime}_{n})p_{+}^{\beta_{\bar{h}}}(g\mid s^{\prime}_{n})ds^{\prime}_{n}} 
~\text{select states correspond to reachable goals by $\hat{f}$} \nonumber\\
&\nonumber\\
&=\int_{s_{n}} \sum_hp(h\mid s,a)p_+^{\beta_h}(s_{n}\mid s,a)\sum_{\tilde{h}}p(\tilde{h}\mid s_{n})p_+^{\beta_{\tilde{h}}}(g\mid s_{n})ds_{n}\pm\epsilon_k L_1\int_{s_{n}} \sum_{\tilde{h}}p(\tilde{h}\mid s_{n})p_+^{\beta_{\tilde{h}}}(g\mid s_{n})ds_{n} \nonumber\\
&=p^{1-\mathrm{step}}(g\mid s,a)\pm\mathcal{O}(\epsilon_k L_1)
&
\end{flalign}
(a) $u$ are the goals are the initial goal clustered by k-means.
And then substituted by its near neighbour $s_n$. 
As $u$ contains $s_n$ and $s^{\prime}_n$,
they follow from Equation \ref{eq:7} and \ref{eq:8} that:
$p_{+}^{\beta_{h}}(u\mid s,a)-\epsilon_k L_1\leq p_{+}^{\beta_{h}}(s_n\mid s,a)\leq p_{+}^{\beta_{h}}(u\mid s,a)+\epsilon_k L_1$
and
$p_{+}^{\beta_{h}}(u\mid s,a)-\epsilon_k L_2\leq p_{+}^{\beta_{h}}(s^{\prime}_n\mid s,a)\leq p_{+}^{\beta_{h}}(u\mid s,a)+\epsilon_k L_2$
\section{Implementation Details} \label{sec: supplementary material_implementation}
\subsection{Datasets}
We utilize the newly proposed offline dataset by \citep{ghugare2024closing} to evaluate the stitching capabilities of GCWSL methods. Specifically, we employ the goal-conditioned point mazes (Umaze, Medium, and Large) as our default datasets. In these mazes, the task involves navigating a ball with two degrees of freedom, actuated by forces in the cartesian x and y directions. Data collection is performed using a PID controller.
\subsection{Hyperparameters}
We conducted a sequential comparison experiment by implementing the GCWSL baselines within a consistent framework. Specifically, all GCWSL baseline implementations are based on the DWSL framework \citep{hejna2023distance}, including the hyperparameters. These hyperparameters were adjusted for specific datasets as it was observed that GCWSL methods can exhibit overfitting with certain combinations of these three values. However, it is expected that the success rate will continue to improve. Detailed hyperparameter configurations for each dataset are provided in Table \ref{tab:alg_hparams}. 
\begin{table}[h]
    \renewcommand\arraystretch{1.2}
    \vspace{-0.5em}
    \centering
    \scalebox{0.95}{
    \begin{tabular}{l|c|c}
    \toprule
    \textbf{Hyperparameters}                  & \textbf{Values} & \textbf{GCWSL-Dataset}                                                          \\
    \hline
    \multirow{2}{*}{Model Architecture}  & $\left[256,256,256\right]$                          & All GCWSL Methods-\texttt{Umaze}; WGCSL-\texttt{Vision-Medium/Large};             \\
                                     & $\left[512,512\right]$                          & Other Combination;                                            \\
    \hline
    \multirow{2}{*}{Batch Size} & 256                          & All GCWSL Methods-\texttt{Umaze}; WGCSL-\texttt{Vision-Medium/Large}; GoFar-\texttt{Large};                                         \\
                                     & 512                          & Other Combination;                                                   \\
    \hline
    \multirow{7}{*}{Total Updates} & 5000 x 4= 20000                          & SMORE-\texttt{Umaze}; WGCSL-\texttt{Vision-Umaze};                                          \\
                                     & 8000 x 3= 24000                          & GoFar-\texttt{Large};                                                   \\
                                     & 5000 x 6= 30000                          & GoFar-\texttt{Umaze}; WGCSL-\texttt{Umaze};                                                  \\
                                     & 5000 x 9= 45000                          & GoFar-\texttt{Medium}; WGCSL-\texttt{Vision-Umaze};                                                   \\
                                     & 5000 x 10= 50000                          & WGCSL-\texttt{Medium}; DWSL-\texttt{Medium}; SMORE-\texttt{Medium};                                                  \\
                                     & 8000 x 8= 64000                          & DWSL-\texttt{Large};\\
                                     & 8000 x 10= 80000                          & Other Combination.\\
    \bottomrule
    \end{tabular}
    }
    \caption{GCWSL methods hyperparameters under specific dataset. Total updates = updates per iters x iters}
    \label{tab:alg_hparams}
\end{table}

The hyperparameters for SGDA and TGDA remain consistent with their original settings. The hyperparameters for MGDA are based on the configuration used in classic robotics environments as described in \citep{ke2024ccil}. To ensure a fair comparison, the same augmentation probability was applied to each augmentation methods. Detailed hyperparameters for these methods can be found in Table \ref{tab:alg_hparams_2}.
\begin{table}[h]
    \renewcommand\arraystretch{1.2}
    \vspace{-0.5em}
    \centering
    \scalebox{0.9}{
    \begin{tabular}{l|c|c}
    \toprule
    \textbf{Algorithms}                  & \makecell[c]{\textbf{Hyperparameters}} & \makecell[c]{\textbf{Value}}                                                          \\
    \hline
    \multirow{1}{*}{SGDA \citep{yang2023swapped}}  & \makecell[c]{$\epsilon$}                          & \makecell[c]{0.5}              \\
    \hline
    \multirow{4}{*}{TGDA \citep{ghugare2024closing}} & \makecell[c]{$C$ of \texttt{Umaze}}                         & \makecell[c]{20}                                         \\
                                     & \makecell[c]{$C$ of \texttt{Medium}}                         & \makecell[c]{40}                                                       \\
                                     & \makecell[c]{$C$ of \texttt{Large}}                         & \makecell[c]{80} \\
                                     & \makecell[c]{$\epsilon$}                         & \makecell[c]{0.5}                                            \\
    \hline
    \multirow{4}{*}{\textbf{MGDA (Ours)}} & \makecell[c]{$\lambda$}                          & \makecell[c]{1.0}                                          \\
                                     & \makecell[c]{$K$}                         & \makecell[c]{2.0}\\
                                     & \makecell[c]{$\epsilon$}                          & \makecell[c]{0.5} \\
                                     & \makecell[c]{$\delta$}                          & \makecell[c]{0.5} 
                                     \\
    \bottomrule
    \end{tabular}}
    \caption{Goal data augmentation works specific hyperparameters. $C$ denotes the number of clusters for k-means. $\epsilon$ is the probability of augmenting a goal.} 
\label{tab:alg_hparams_2}
\end{table}
\section{The Necessity of MGDA for GCWSL Methods}
\begin{figure*}[!h]
    \centering
    \begin{minipage}{\linewidth}
		\vspace{3pt}
		\centerline{\includegraphics[width=0.4\textwidth]{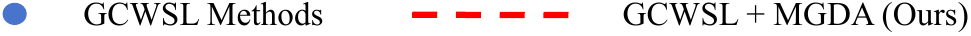}}
    \end{minipage}
    \begin{minipage}{0.33\linewidth}
		\vspace{3pt}
		\centerline{\includegraphics[width=0.92\textwidth]{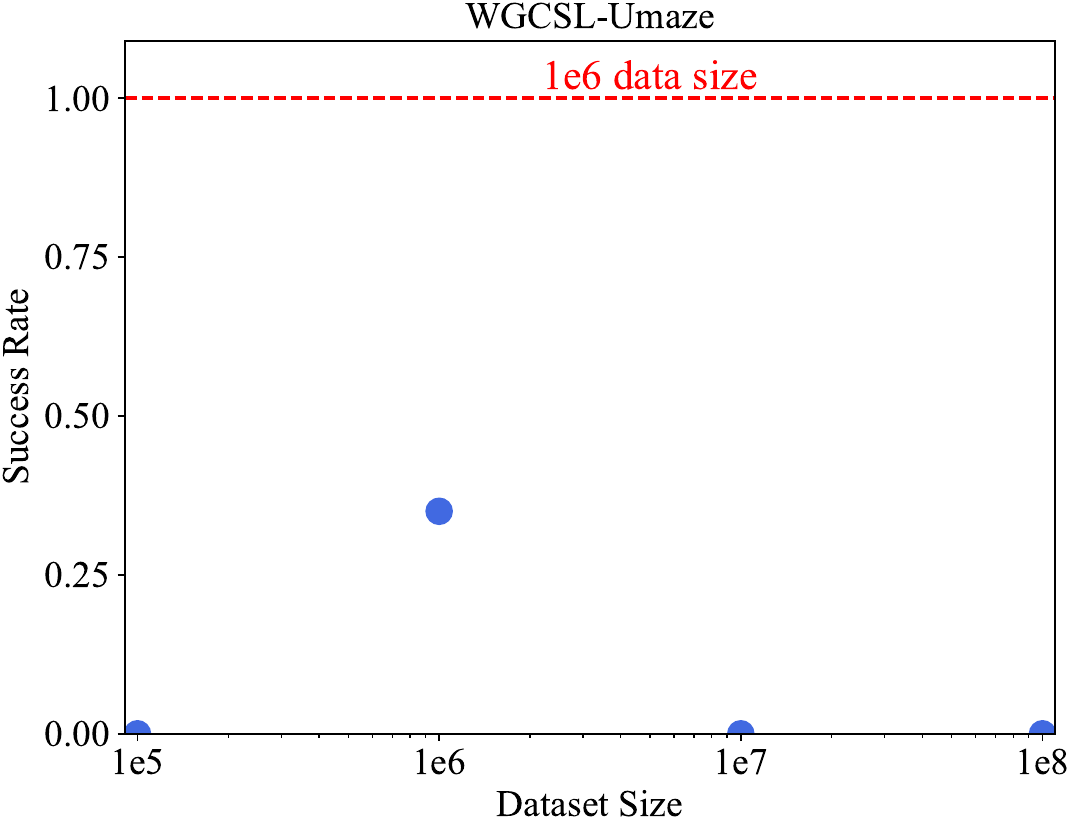}}
	\end{minipage}
	\begin{minipage}{0.33\linewidth}
		\vspace{3pt}
		\centerline{\includegraphics[width=0.92\textwidth]{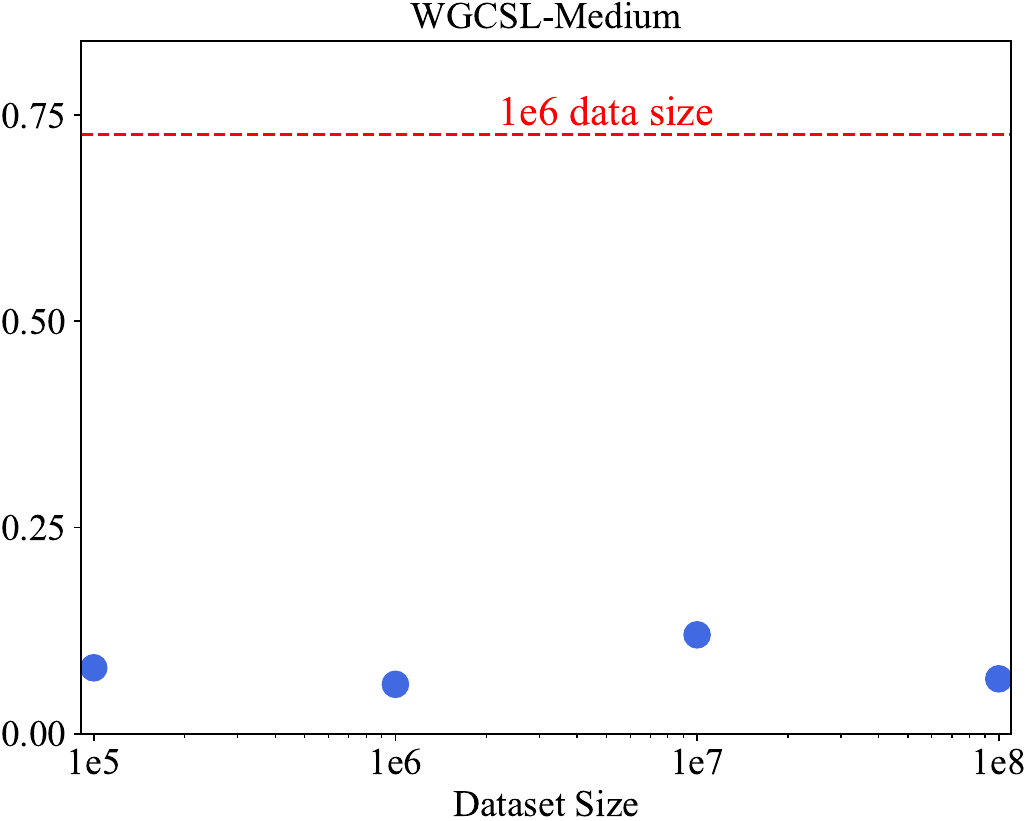}}
	\end{minipage}
    \begin{minipage}{0.33\linewidth}
		\vspace{3pt}
		\centerline{\includegraphics[width=0.92\textwidth]{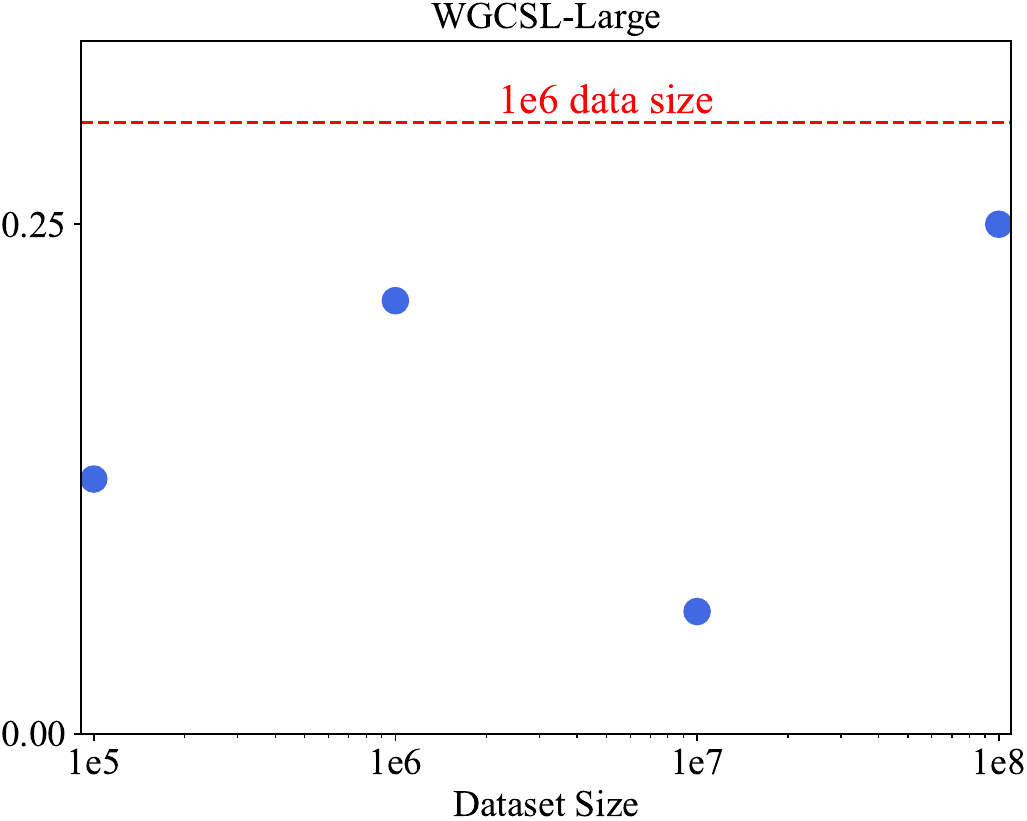}}
	\end{minipage}
    \begin{minipage}{0.33\linewidth}
		\vspace{3pt}
		\centerline{\includegraphics[width=0.92\textwidth]{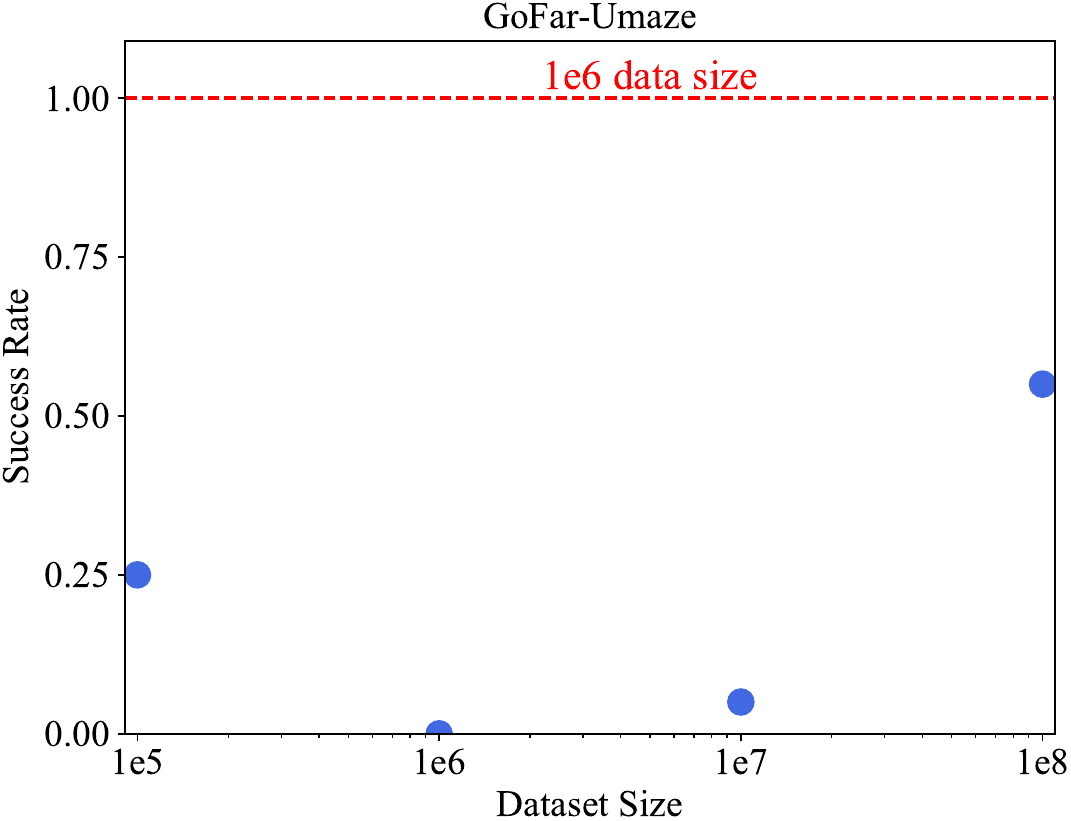}}
	\end{minipage}
	\begin{minipage}{0.33\linewidth}
		\vspace{3pt}
		\centerline{\includegraphics[width=0.92\textwidth]{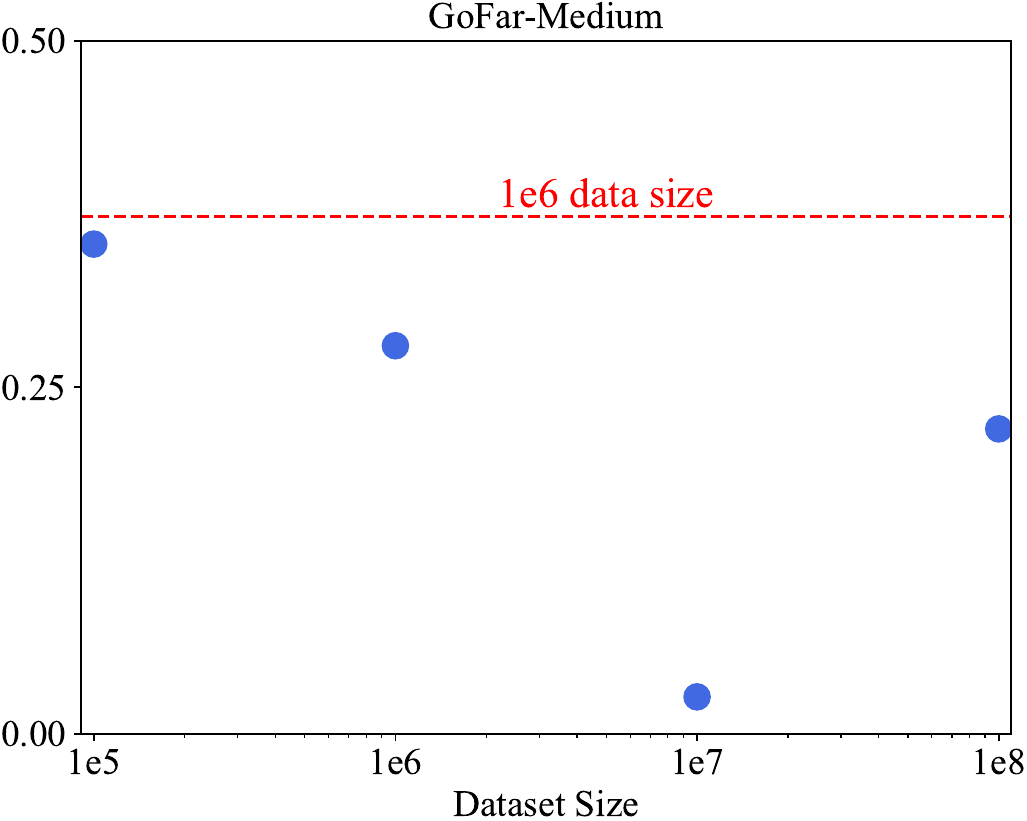}}
	\end{minipage}
    \begin{minipage}{0.33\linewidth}
		\vspace{3pt}
		\centerline{\includegraphics[width=0.92\textwidth]{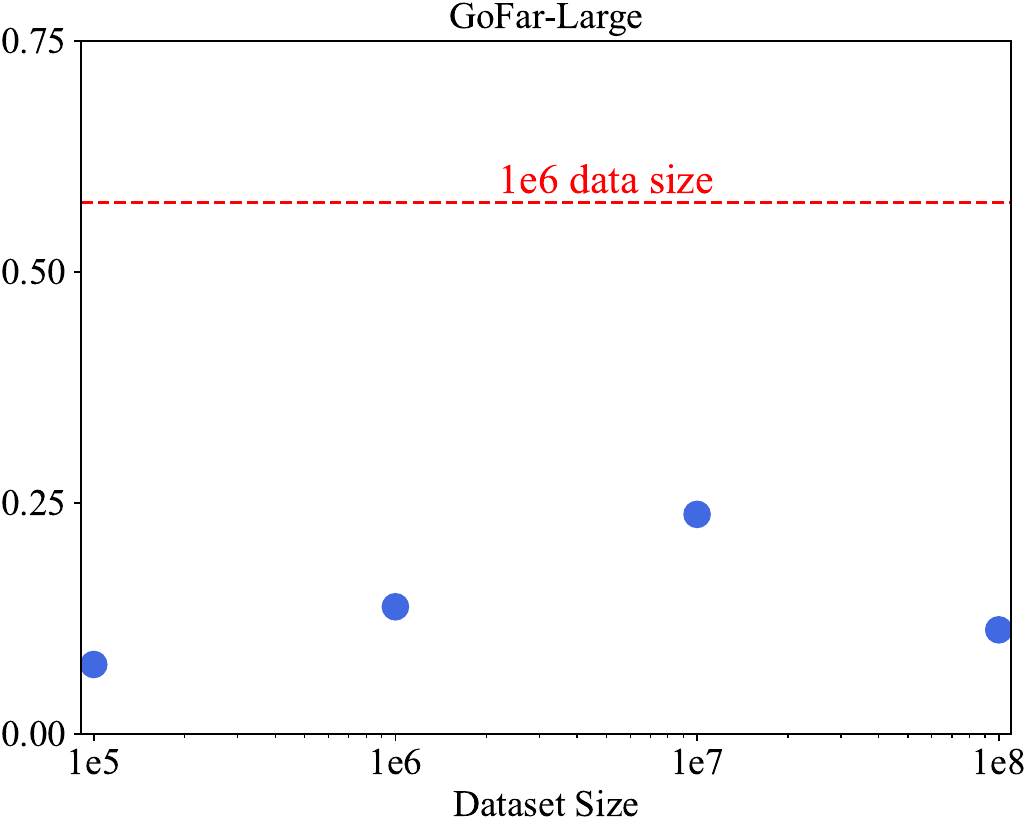}}
	\end{minipage}
    \begin{minipage}{0.33\linewidth}
		\vspace{3pt}
		\centerline{\includegraphics[width=0.92\textwidth]{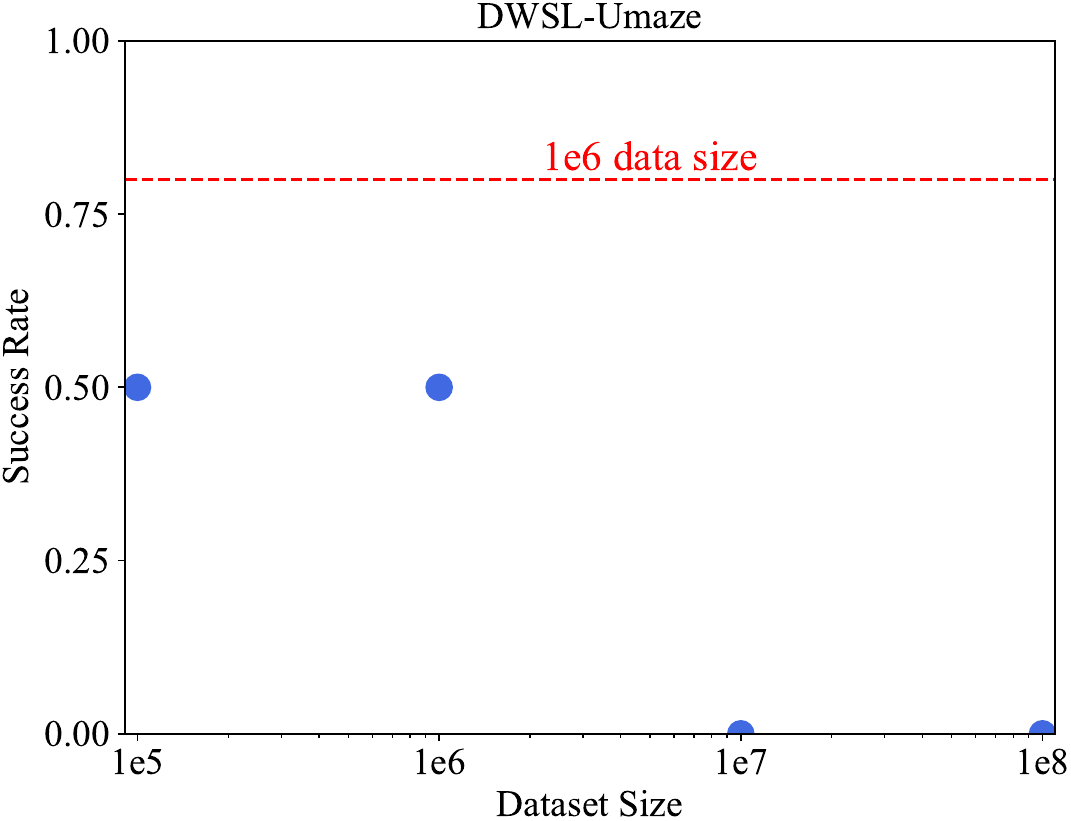}}
	\end{minipage}
	\begin{minipage}{0.33\linewidth}
		\vspace{3pt}
		\centerline{\includegraphics[width=0.92\textwidth]{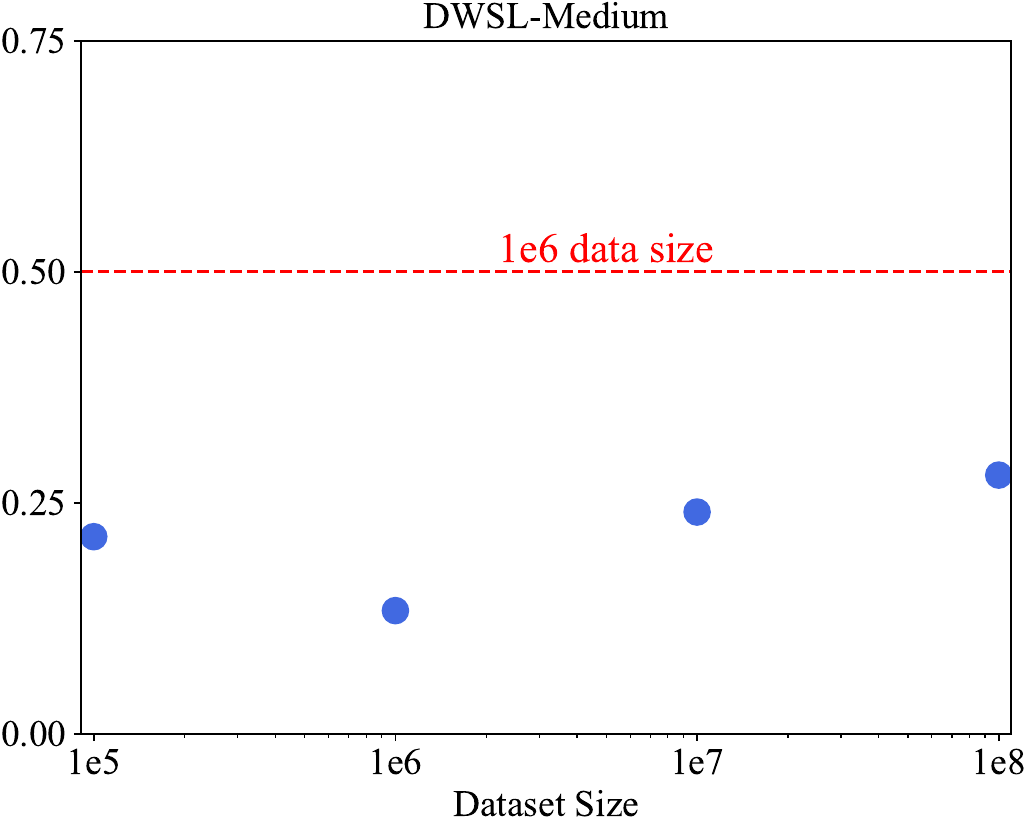}}
	\end{minipage}
    \begin{minipage}{0.33\linewidth}
		\vspace{3pt}
		\centerline{\includegraphics[width=0.92\textwidth]{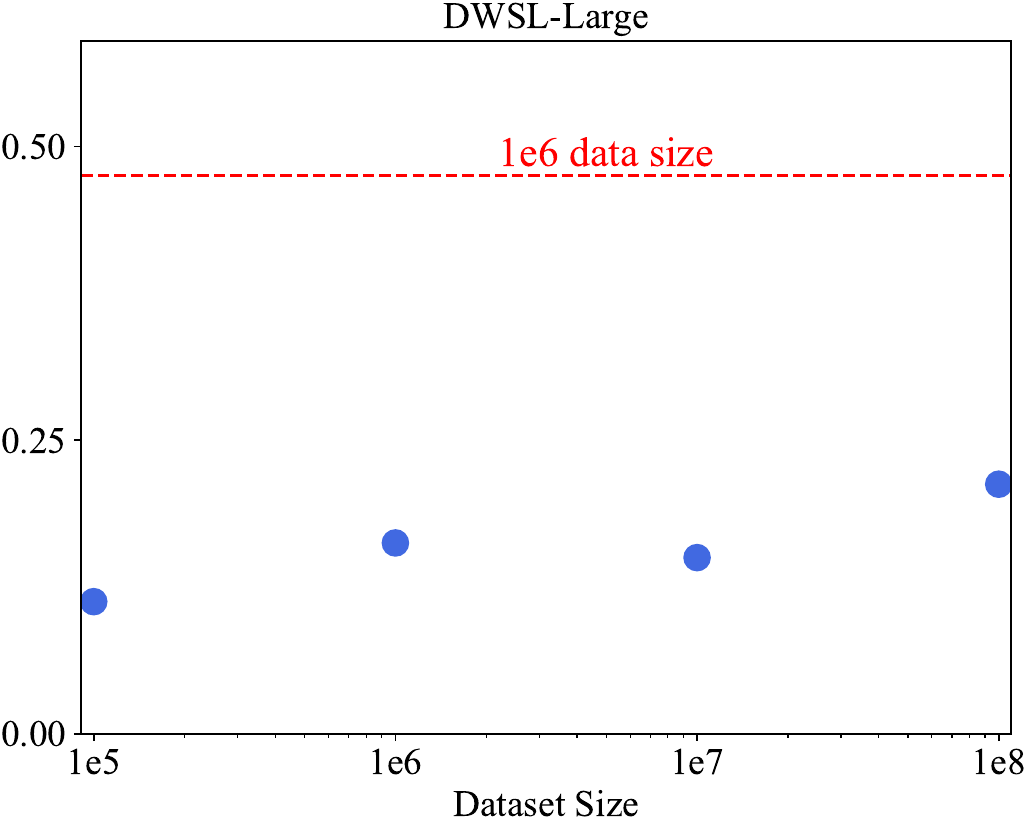}}
	\end{minipage}
    \begin{minipage}{0.33\linewidth}
		\vspace{3pt}
		\centerline{\includegraphics[width=0.92\textwidth]{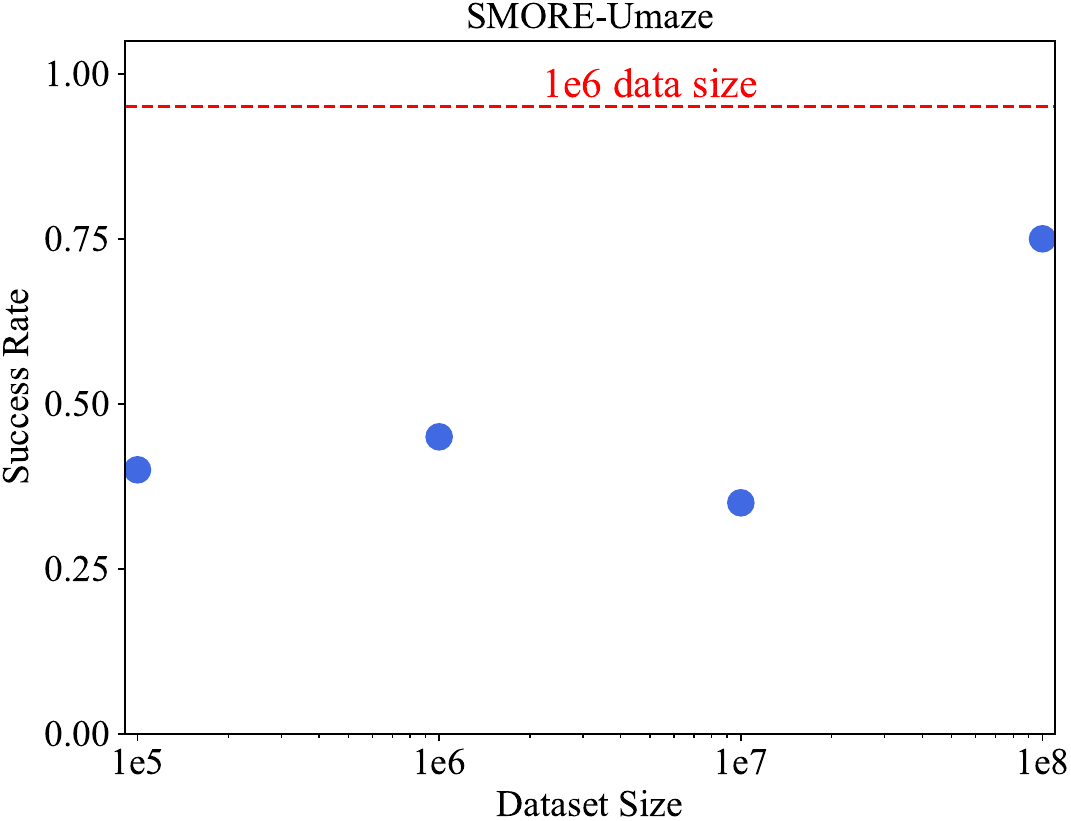}}
	\end{minipage}
	\begin{minipage}{0.33\linewidth}
		\vspace{3pt}
		\centerline{\includegraphics[width=0.92\textwidth]{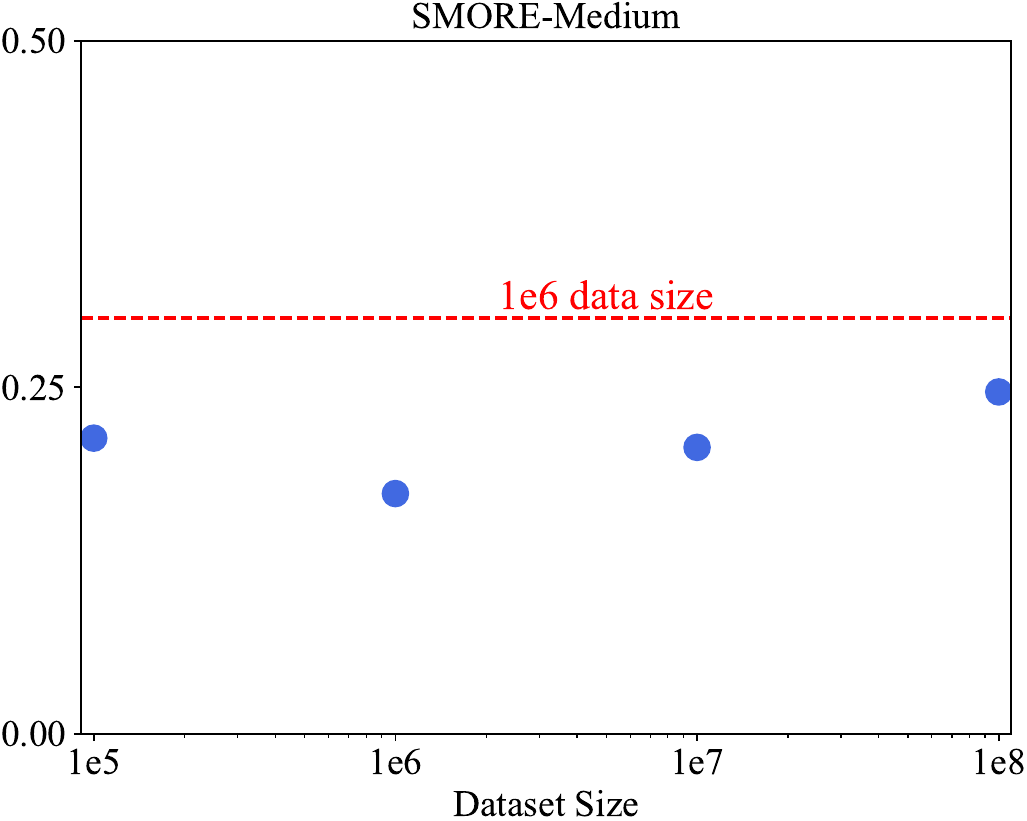}}
	\end{minipage}
    \begin{minipage}{0.33\linewidth}
		\vspace{3pt}
		\centerline{\includegraphics[width=0.92\textwidth]{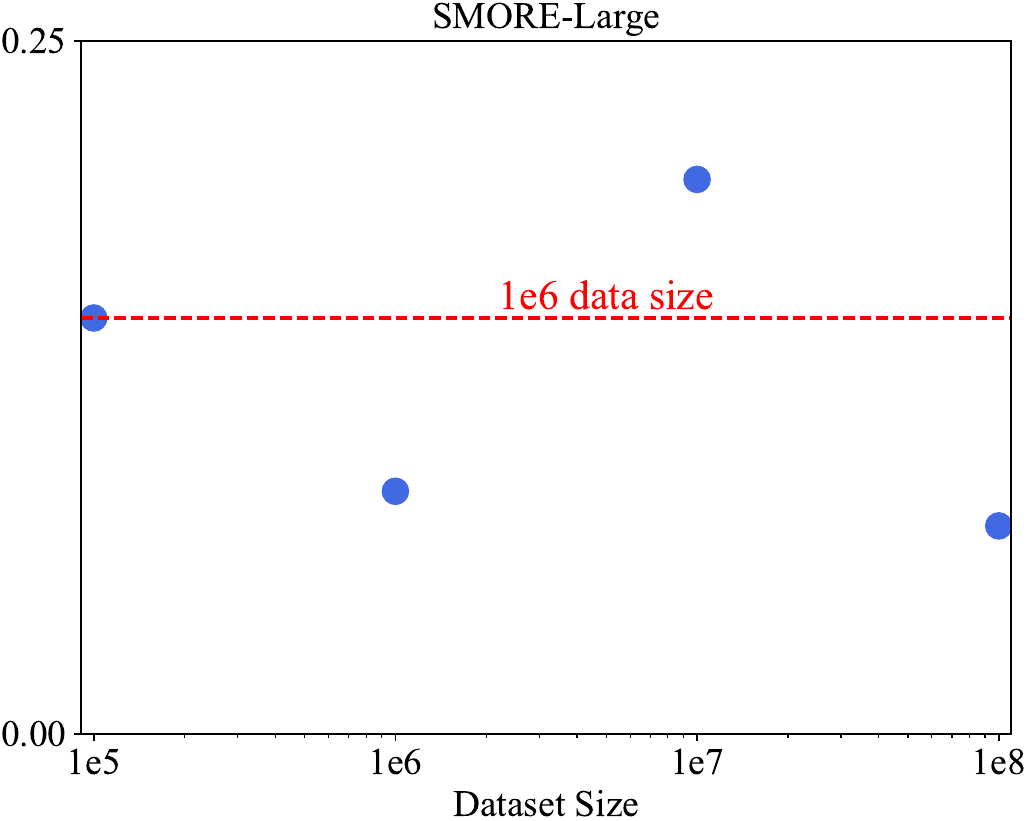}}
	\end{minipage}
    \caption{
    Performance of GCWSL methods trained on three different offline dataset sizes, averaged across all point mazes. Despite the use of larger datasets, the generalization of GCWSL remains inferior to that of GCWSL combined with MGDA.
    }
    \label{fig:smore necessity results}
\end{figure*}

To determine whether additional data can expose more state-goal pairs to GCWSL methods and enhance GCWSL's generalization, this section examines the necessity of data augmentation under specific dataset size constraints. We constructed four datasets of varying sizes for each dataset and algorithm. As shown in Figure \ref{fig:smore necessity results}, applying data augmentation to GCWSL methods with limited dataset sizes generally outperforms other approaches across most datasets.

\end{document}